\documentclass[10pt,twocolumn,letterpaper]{article}

\usepackage{cvpr}
\usepackage{times}
\usepackage{epsfig}
\usepackage{graphicx}
\usepackage{amsmath}
\usepackage{amssymb}

\usepackage[pagebackref=true,breaklinks=true,letterpaper=true,colorlinks,bookmarks=false]{hyperref}

\usepackage{bm} %
\usepackage{subfig}
\usepackage{alphalph} %
\usepackage{booktabs}  %

\usepackage[linesnumbered,ruled]{algorithm2e} %

\DeclareMathOperator*{\argmin}{argmin}

\cvprfinalcopy %

\def\cvprPaperID{031} %

\ifcvprfinal\fi
\begin{document}

\title{Deep Face Deblurring}

\author{Grigorios G. Chrysos \\
Imperial College London \\
{\tt\small g.chrysos@imperial.ac.uk}
\and
Stefanos Zafeiriou \\
Imperial College London \\
{\tt\small s.zafeiriou@imperial.ac.uk}
}

\maketitle
\begin{abstract}
Blind deblurring consists a long studied task, however the outcomes of generic methods are not effective in real world blurred images. Domain-specific methods for deblurring targeted object categories,  \eg text or faces, frequently outperform their generic counterparts, hence they are attracting an increasing amount of attention. 
In this work, we develop such a domain-specific method to tackle deblurring of human faces, henceforth referred to as face deblurring. Studying faces is of tremendous significance in computer vision, however face deblurring has yet to demonstrate some convincing results. This can be partly attributed to the combination of i) poor texture and ii) highly structure shape that yield the contour/gradient priors (that are typically used) sub-optimal. In our work instead of making assumptions over the prior, we adopt a learning approach by inserting weak supervision that exploits the well-documented structure of the face. Namely, we utilise a deep network to perform the deblurring and employ a face alignment technique to pre-process each face. We additionally surpass the requirement of the deep network for thousands training samples, by introducing an efficient framework that allows the generation of a large dataset. We utilised this framework to create $2MF^2$, a dataset of over two million frames. We conducted experiments with real world blurred facial images and report that our method returns a result close to the sharp natural latent image.
\end{abstract}

\section{Introduction}
\label{sec:deblurring_introduction}

Blind deblurring is the task of acquiring an estimate of the sharp latent image given a blurry image as input. 
No single algorithm for deblurring all objects exists; the task is notoriously ill-posed. To that end, methods that exploit domains-specific knowledge have emerged for deblurring targeted categories of objects, \eg text or faces. Similarly, the focus of this work is face deblurring; we argue that exploiting domain-specific knowledge can lead to superior deblurring results, especially for the human face that presents a highly structured shape. 
Despite the fact that the human face is among the most studied objects in computer vision with significant applications in face recognition, computer graphics and surveillance, face deblurring has not received much attention yet. 

\begin{figure*}[!htb]
    \centering
    \includegraphics[width=0.13\linewidth]{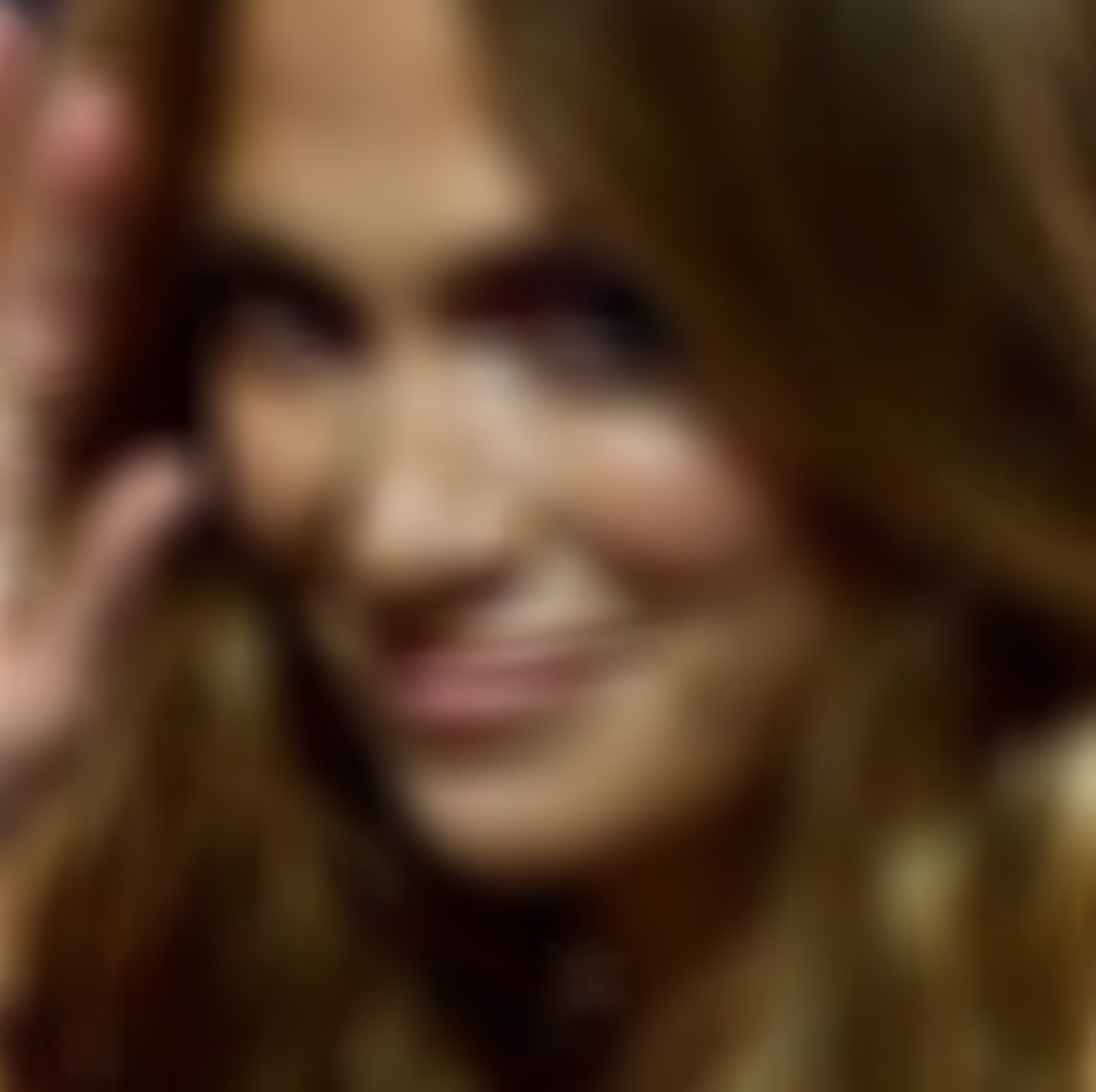}
    \includegraphics[width=0.13\linewidth]{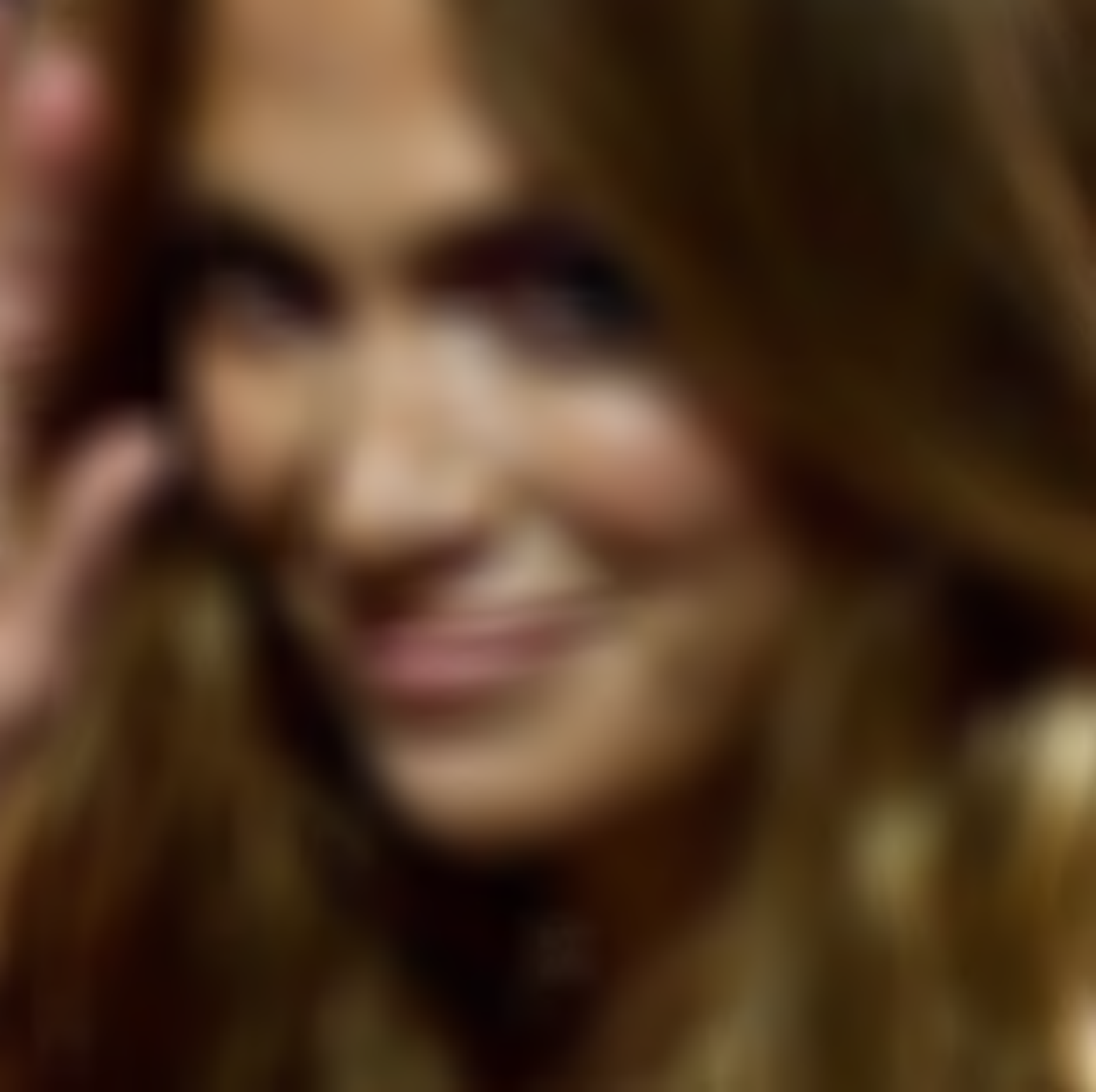}
    \includegraphics[width=0.13\linewidth]{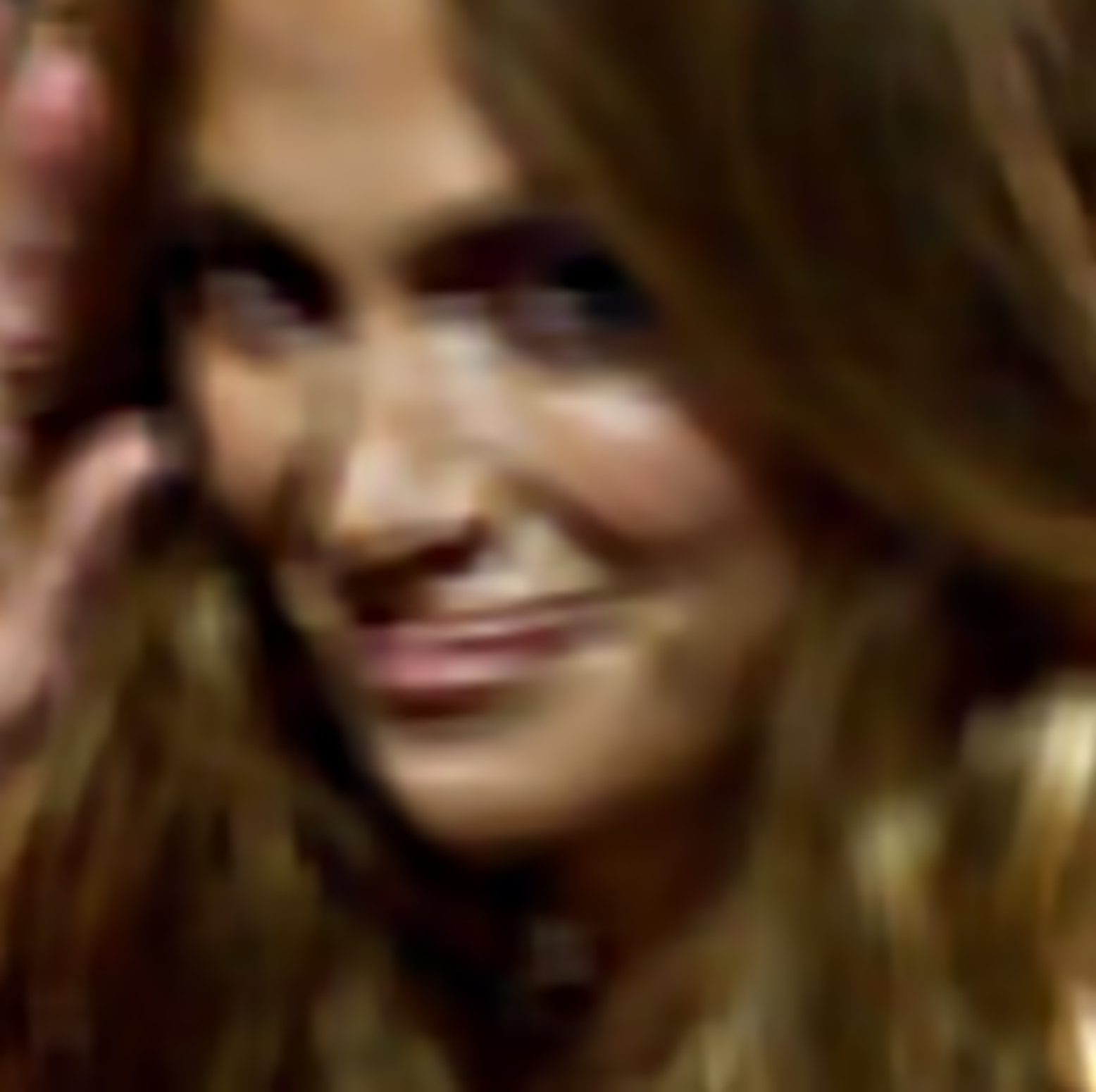}
    \includegraphics[width=0.13\linewidth]{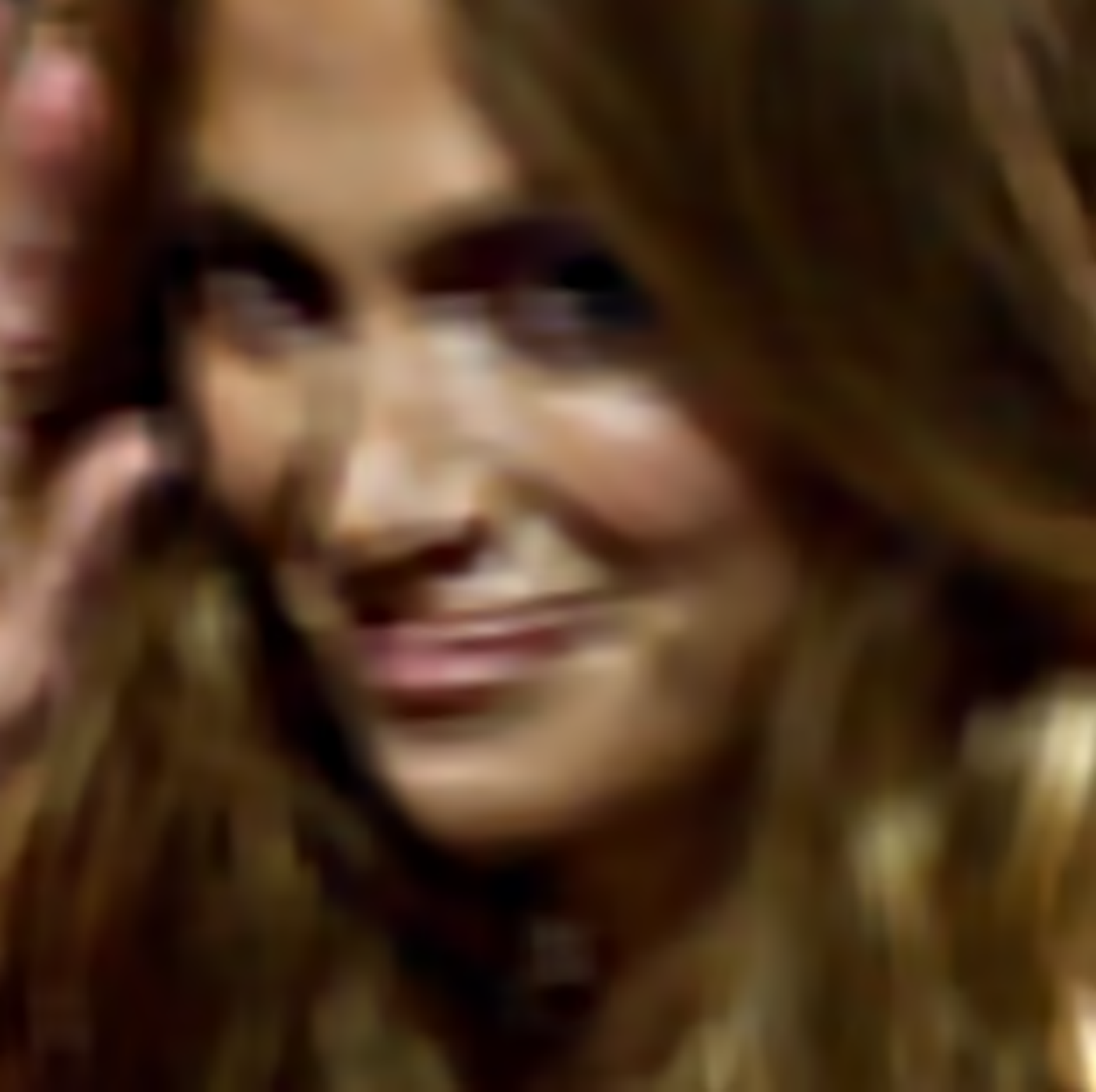}
    \includegraphics[width=0.13\linewidth]{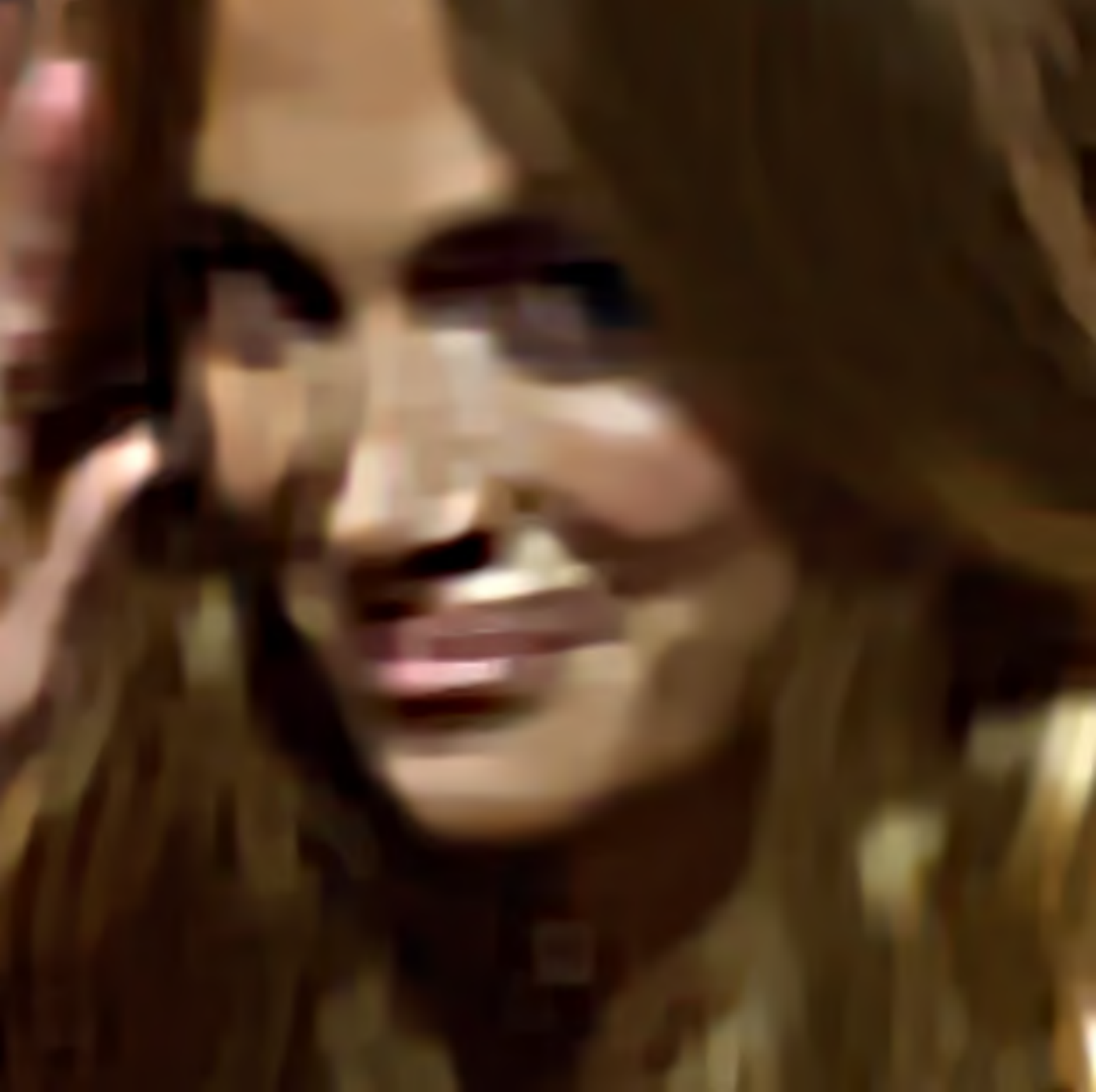}
    \includegraphics[width=0.13\linewidth]{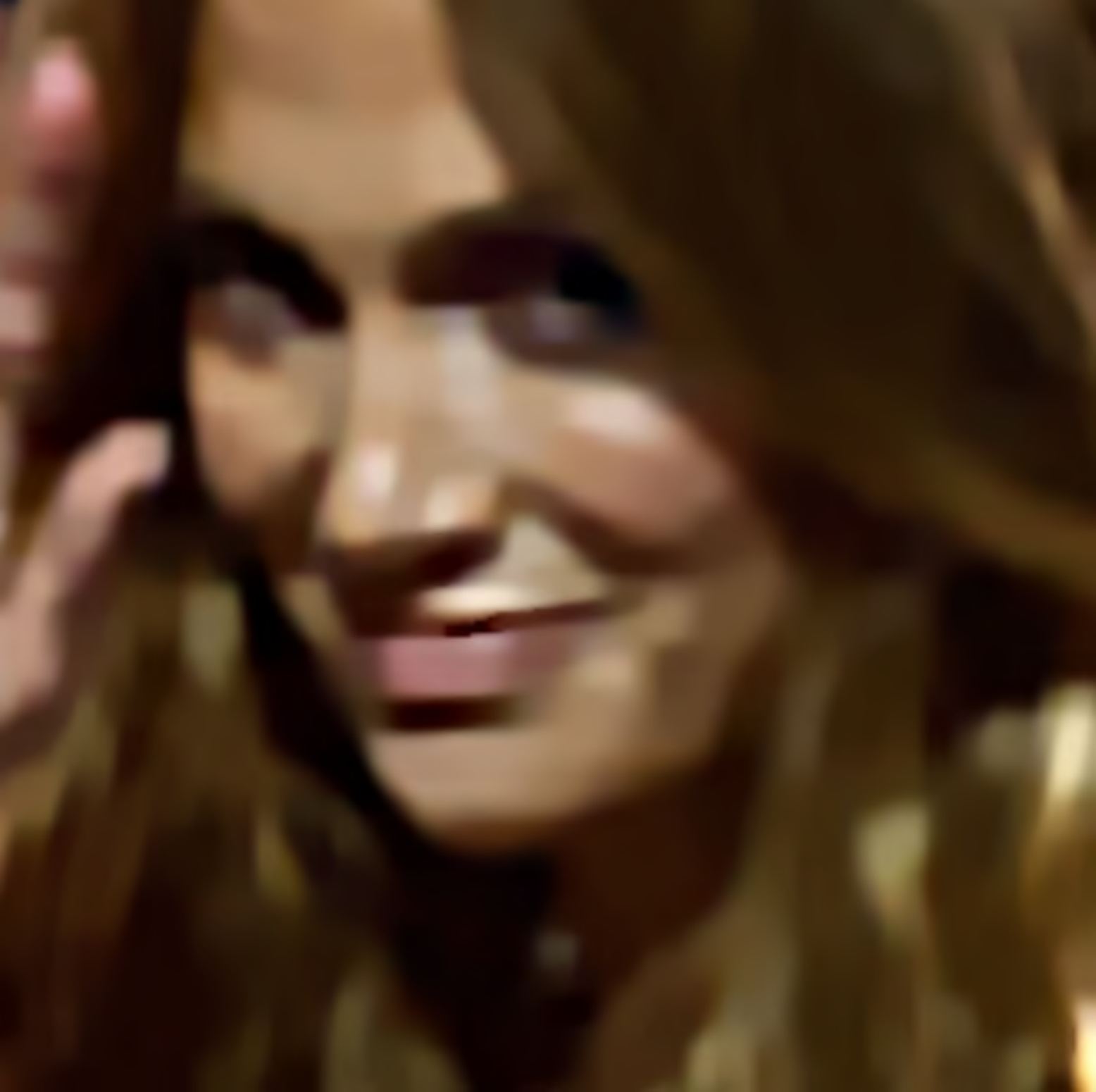}
    \includegraphics[width=0.13\linewidth]{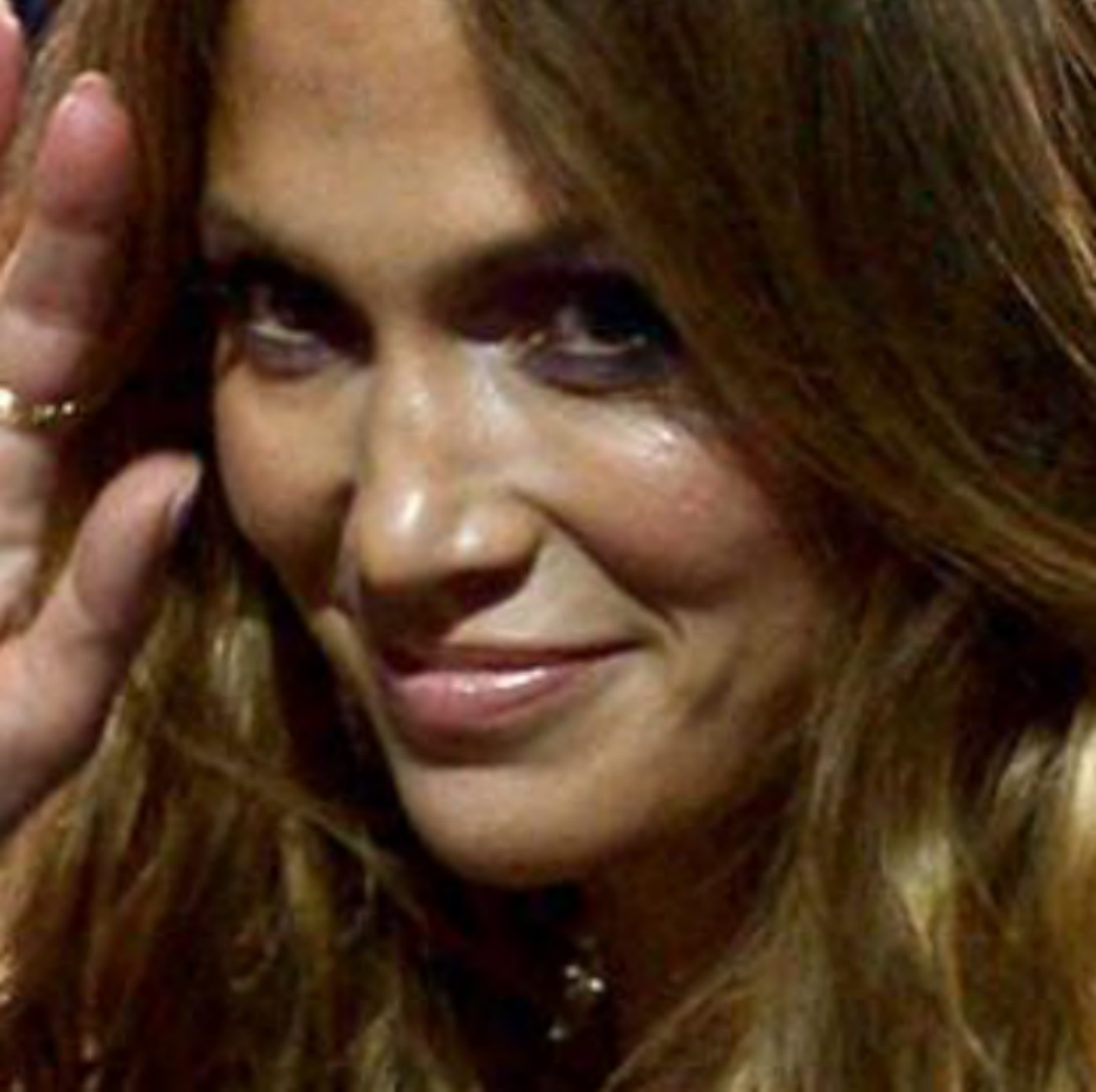}
    \\ \vspace{-2mm}
    \subfloat[][Blurred image]{
    \includegraphics[width=0.13\linewidth]{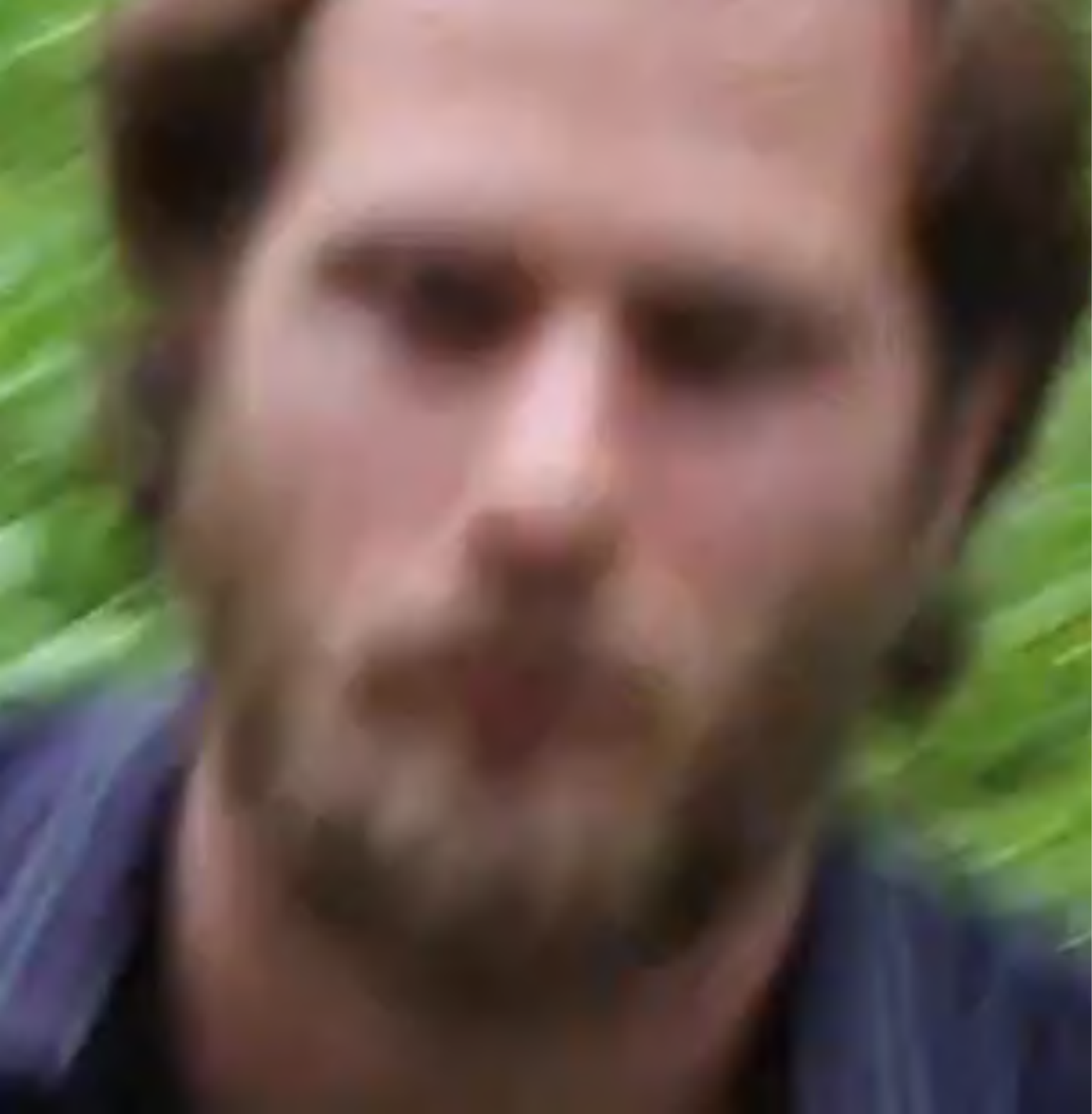}}
    \subfloat[][Babacan \etal~\cite{babacan2012bayesian}]{ \includegraphics[width=0.13\linewidth]{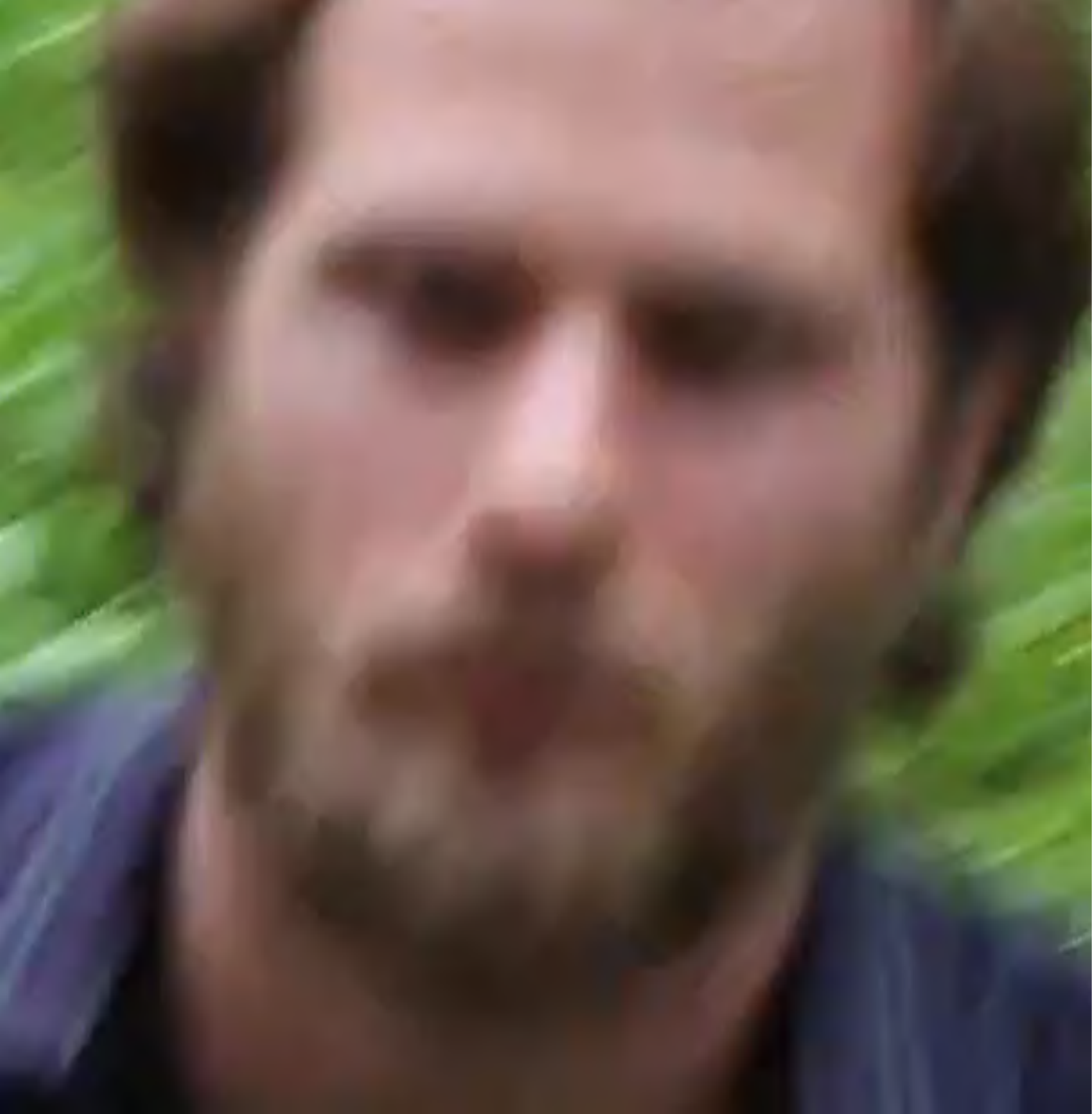}}
    \subfloat[][Zhang \etal~\cite{zhang2013multi}]{
    \includegraphics[width=0.13\linewidth]{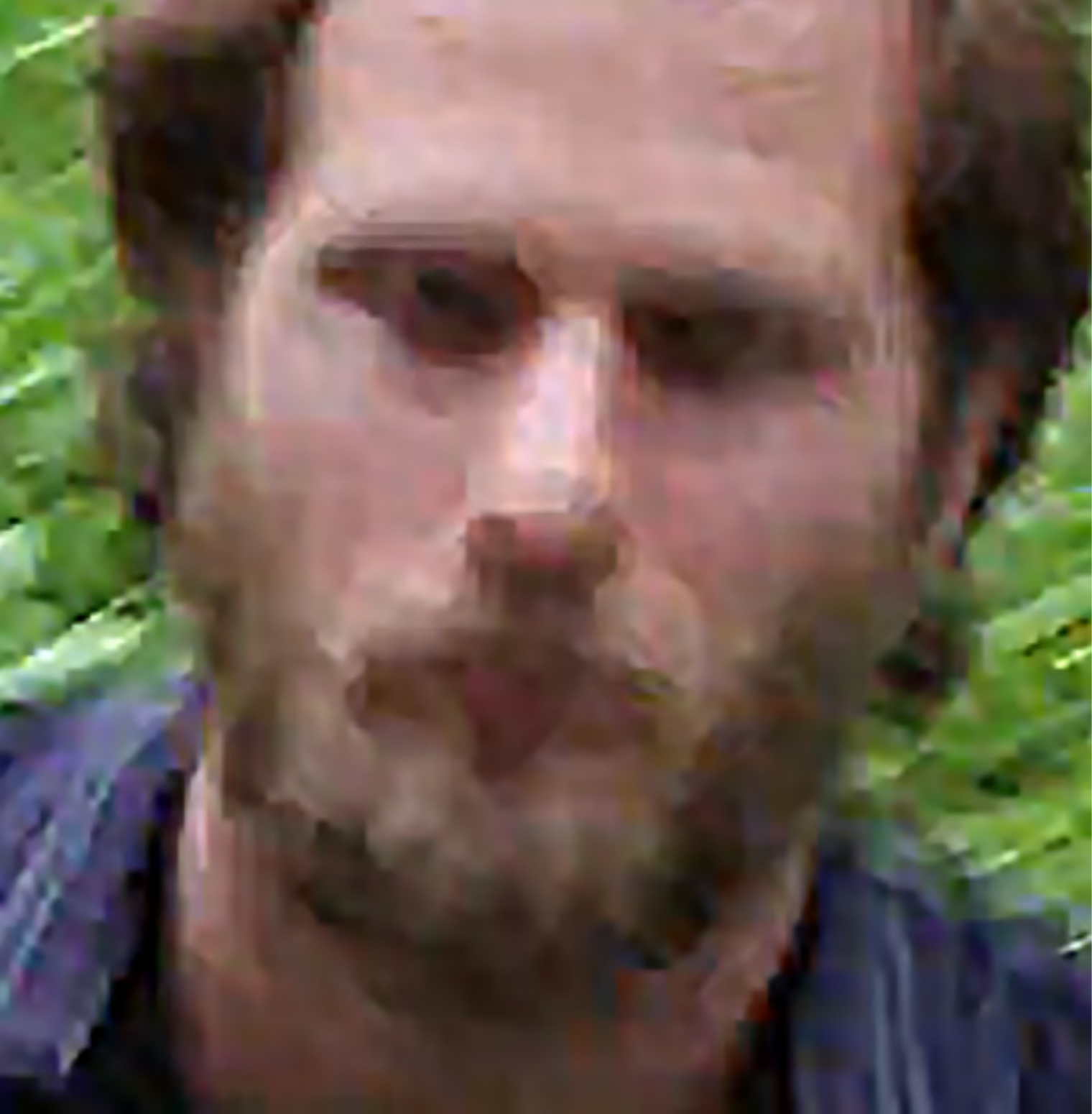}}
    \subfloat[][Pan \etal~\cite{pan2014deblurring}]{
    \includegraphics[width=0.13\linewidth]{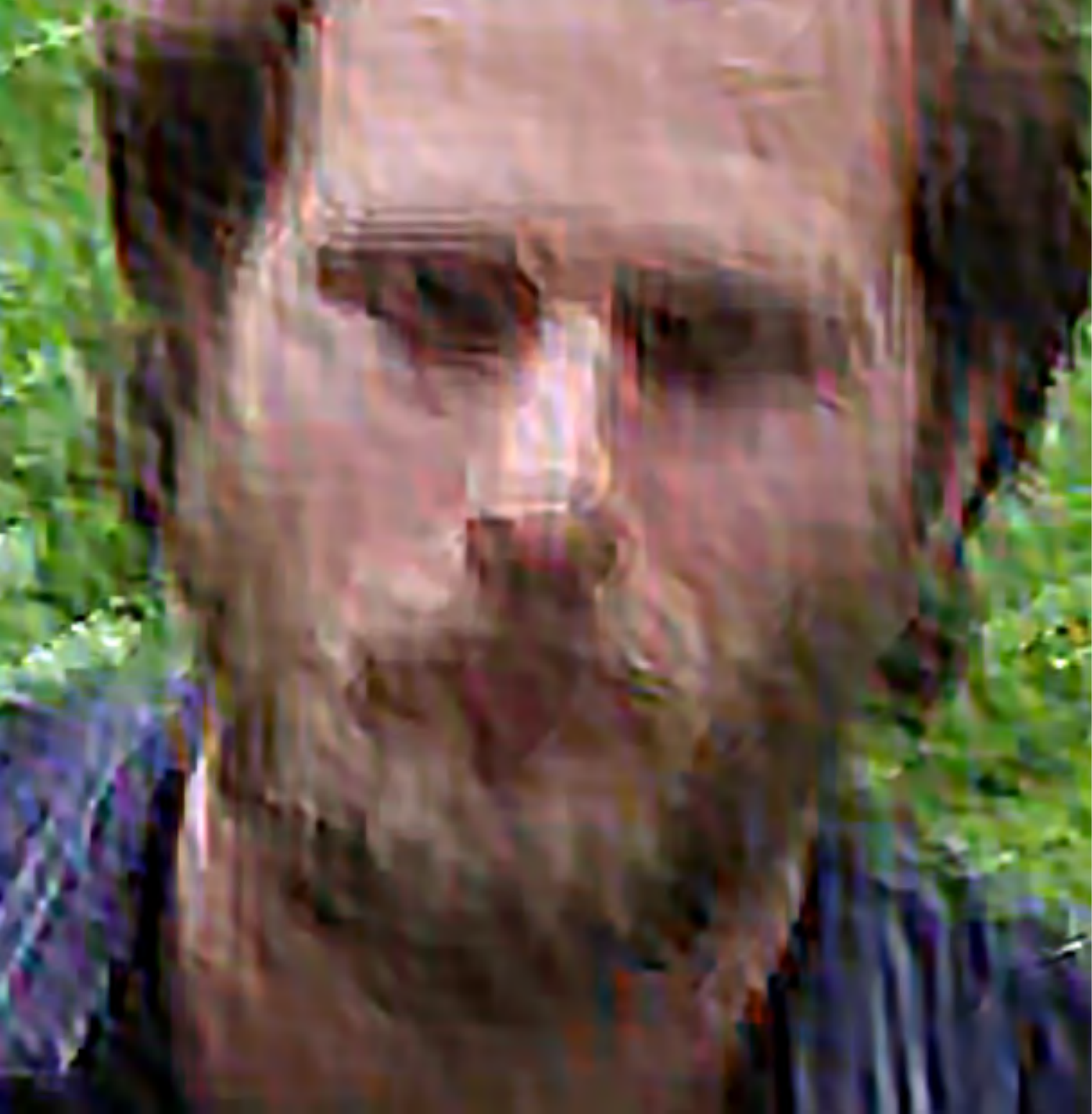}}
    \subfloat[][Pan \etal~\cite{pan2014deblurring_text}]{
    \includegraphics[width=0.13\linewidth]{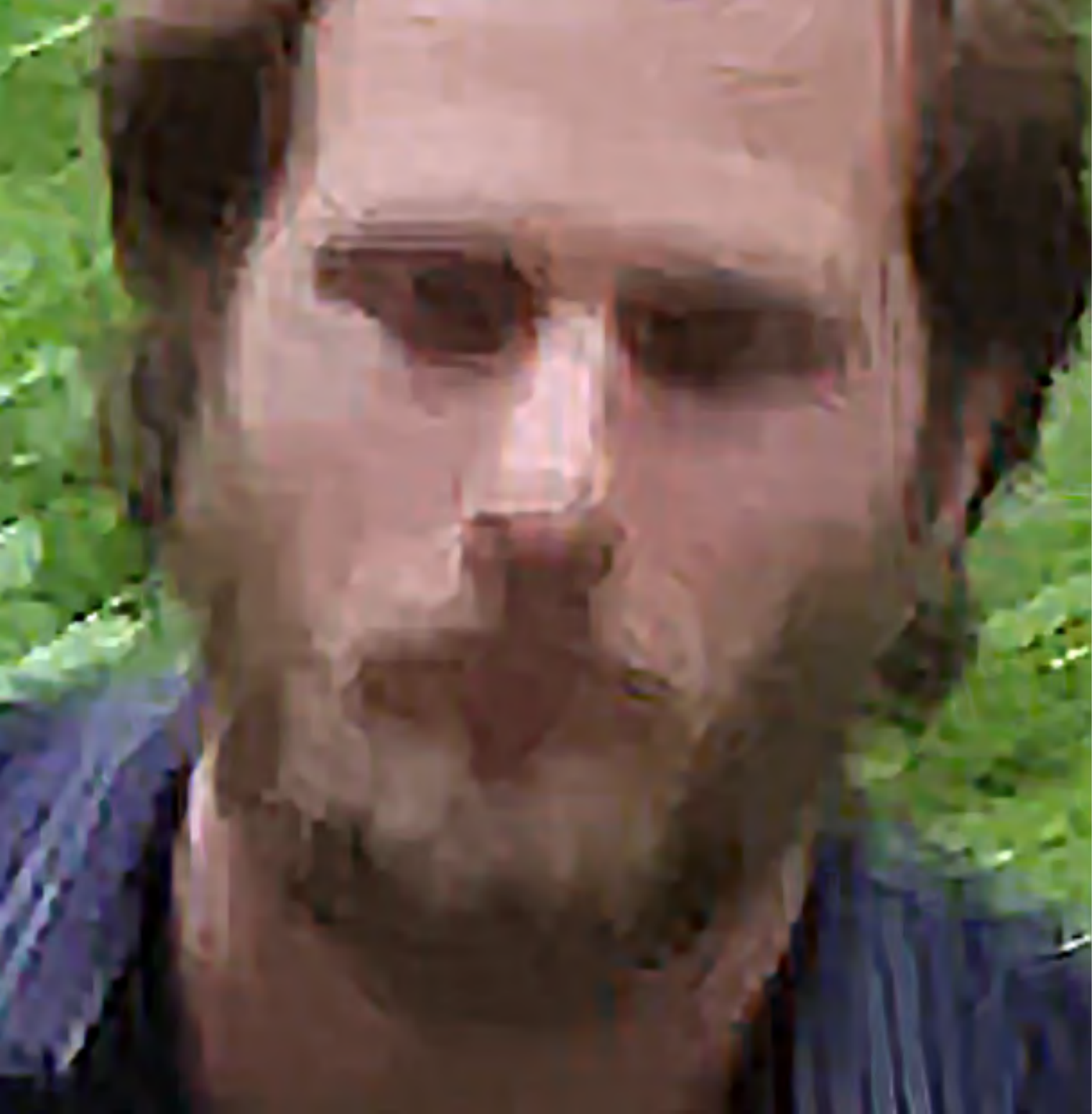}}
    \subfloat[][Pan \etal~\cite{panblind}]{
    \includegraphics[width=0.13\linewidth]{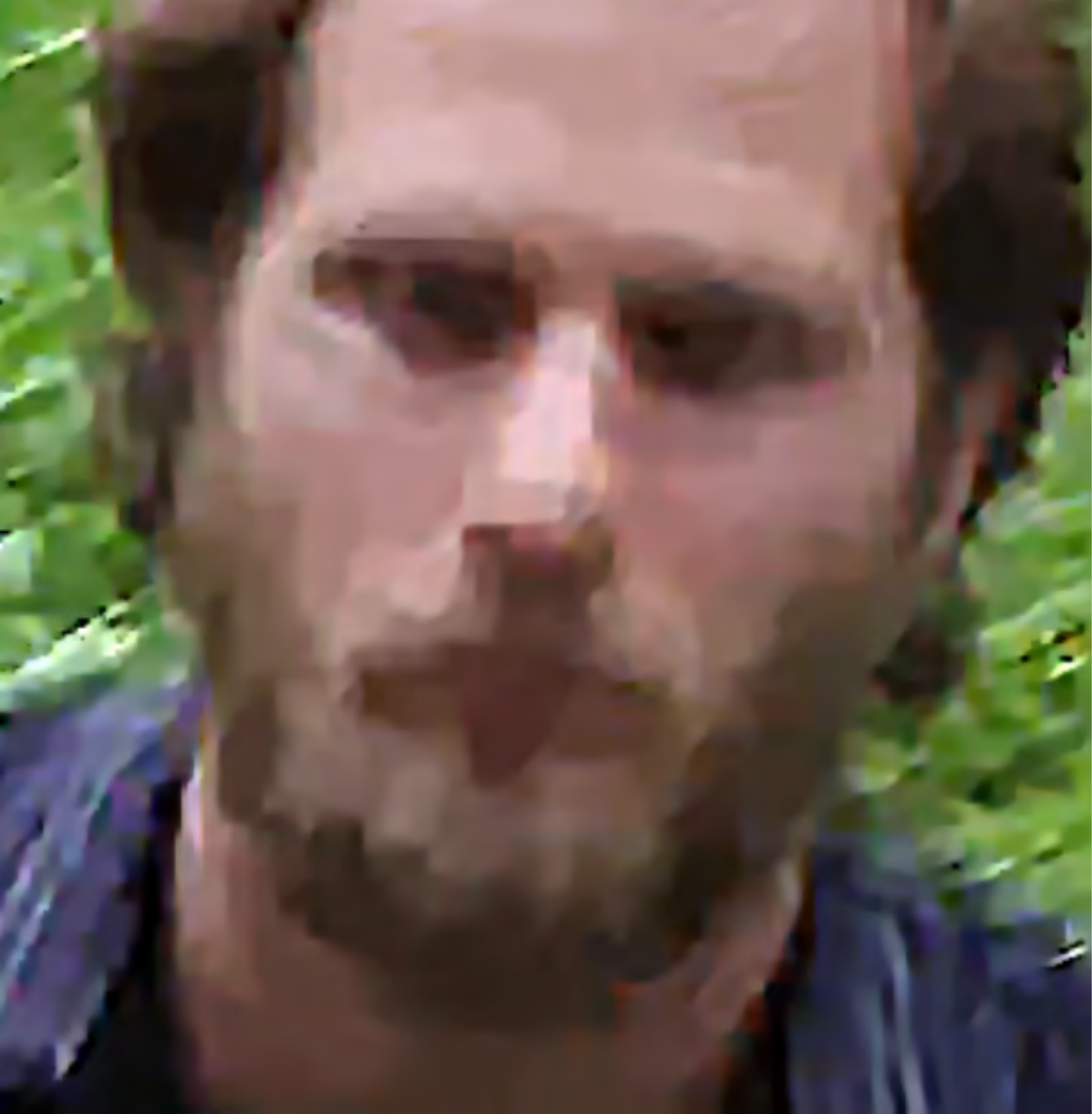}}
    \subfloat[][Original image]{
    \includegraphics[width=0.13\linewidth]{figures/deblurring_faces/motivational_fig/blurry_real.pdf}}
    \vspace{-0.5mm}
    \caption{(Preferably viewed in colour) Two sample facial images as deblurred by the existing methods. The one on the top row was synthetically blurred with a uniform kernel, while the one on the bottom is a real world blurred image. Evidently, the existing methods do not yield a sharp natural facial image as we would expect. The difference between deblurring results depending on the type of blur as emphasized in \cite{laicomparative} can be visually confirmed.}
    \label{fig:sample_deblurring_liter}
\end{figure*}

Deblurring has long been studied (\cite{tekalp1986identification, cho1991blur, levin2009understanding, pan2014deblurring, panblind}), however the results are far from satisfactory (\cite{laicomparative}) when it comes to real world blurred images. As illustrated in Fig.~\ref{fig:sample_deblurring_liter} the result from state-of-the-art methods in real world blurred images (row 2) is far worse than the synthetically blurred images (row 1).
The difficulty in real world blurred images can be attributed to the non-linear functions involved in the imaging process, like lens saturation, depth variation, lossy compression.
Nevertheless, optimisation-based deblurring techniques (\cite{pan2014deblurring_text, levin2009understanding, hacohen2013deblurring, panblind}) have reported some progress, credited to a meticulous choice of priors along with some optimisation restrictions (\cite{levin2009understanding, perrone2014total}). 
Apart from the generic deblurring methods which are applied to all objects (\cite{levin2009understanding, hacohen2013deblurring, perrone2014total}), there are also methods that utilise domain-specific knowledge, \eg text or face priors (\cite{pan2014deblurring_text, pan2014deblurring}). Domain-specific methods frequently outperform their generic counterparts due to their stronger form of supervision. 

The human face includes some characteristics, \eg fairly restricted shape, that allow a stronger form of supervision.
To the best of our knowledge, the  method of \cite{pan2014deblurring} is currently the only method that explicitly models the blurring for the human face. The authors' motivation relies in capitalising on the restricted facial shape to guide their optimisation. Their method computes the external contour of the face and matches it with an exemplar image; then the contour of the exemplar match is used as a prior.
The contour matching restricts the usage of the method since a) it is computationally demanding to compare each image against a dataset, b) the matching is inaccurate for poses that do not exist in the dataset. 
In contrast to \cite{pan2014deblurring}, most of the generic methods yield sub-optimal results in face deblurring, since they include either a prior based on the gradient or a contour/edge detection step. The highly structured facial shape along with the poor texture constitute the reasons why those generic methods are sub-optimal. 

In our work, instead of `intuitively' adding priors or making other assumptions, we embrace a learning-based approach (by capitalising on the recent developments on Convolutional Neural Networks (CNN)), guided by a weak supervision to express the restriction for the shape structure. 
The last few years the introduction of elaborate benchmarks~\cite{russakovsky2015imagenet} allowed CNN methods to surpass the performance of the hand-crafted linear optimisation techniques, \eg in detection~\cite{girshick2015fast, bell2015ion}, model-free tracking~\cite{nam2016mdnet}, classification~\cite{he2015deep}. The core component of our architecture is the state-of-the-art residual network (ResNet), which is discriminatively trained from training samples of sharp/blurry facial images. A form of weak supervision is introduced by aligning predefined landmark points in the face. This pre-processing step allows the network to encapsulate our restriction for a particular facial structure. We do not enforce a strict alignment as often performed in the landmark localisation techniques (\cite{kazemi2014one, zhu2015face}), since warping creates non-linear artifacts.
Additionally, the blurring process might lead to an ambiguity in the exact positioning of the landmarks, hence deblurring might not be as trivial in case of strict alignment. However, in our experimentation the localisation works sufficiently well for our purpose of selecting and pre-processing the region to be fed in the network.

A constraint of the (supervised) learning-based methods is their dependency on massive amounts of training samples. Collecting and annotating such datasets (\cite{russakovsky2015imagenet, charles2014automatic, zhang2013actemes, shen2015first}) is expensive and laborious, hence there is an increasing effort to create datasets semi-automatically~\cite{shen2015first, sagonas2015300} or almost in an unsupervised manner~\cite{charles2014automatic}. We rectify that for our task by devising an automatic framework that allows the creation of a large dataset with human faces from videos. The framework can select the appropriate frames completely automatically, however in our case a user verified that a face is included in the last frame of each video.
We have utilised this framework to create $2MF^2$, a dataset with millions of facial frames.  $2MF^2$ consists of over a thousand video clips with an accumulated number of 2,1 million frames, which constitutes $2MF^2$ the largest dataset of video frames for faces\footnote{The alternatives of 300VW~\cite{shen2015first} and Youtube Faces~\cite{wolf2011face} include 250 and 620 thousand frames respectively. Furthermore, the Youtube Faces is not appropriate for discriminative learning, since many of the clips are already blurred and of low resolution.}.

Our contributions can be summarised as: 
\begin{itemize}
	\item We introduce a network architecture that performs face deblurring. We validated the trained model in different experiments including synthetically blurred images, images with simulated motion blur as well as low resolution real world blurred images.
	\item We introduce an automatic framework that allows the collection of large datasets in a time-efficient manner. We have utilised this framework to create the $2MF^2$ dataset, which consists of more than 2 million frames. 
\end{itemize}

In the following Sections we summarise the related methods in Sec.~\ref{sec:deblurring_related}; develop our method in Sec.~\ref{sec:deblurring_method}; describe the framework we have devised in Sec.~\ref{sec:deblurring_dataset} and finally experimentally validate our method in Sec.~\ref{sec:deblurring_experiments}.
\section{Related Work}
\label{sec:deblurring_related}

Blur is typically modelled as the convolution of a blur kernel $\bm{K}_g$ with a (latent) sharp image $\bm{I}$, i.e. 
\begin{equation}
\label{equ:main_blurring_model}
    \bm{I}_{bl} = \psi(\bm{I} * \bm{K}_g + \bm{\epsilon})
\end{equation}
with $\bm{I} \in \mathbb{R}^{h_1 \times w_1}$, $\bm{K}_g \in \mathbb{R}^{h_2 \times w_2}$ ($h_2 \ll h_1$, $w_2 \ll w_1$),  
while $\bm{I}_{bl} \in \mathbb{R}^{a \times b}$ denotes the blurry image with $a = h_1 - h_2 + dh, b = w_1 - w_2 + dw$. The $dh, dw$ depend on the type of convolution (`$*$') chosen. The symbol of $\bm{\epsilon}$ represents the noise term, $\psi$ a function that models additional non-linear artifacts, \eg lossy compression, saturated regions, non-linear sensor response.
Retrieving the sharp image is an ill-posed problem, thus some strong assumptions/priors are required. A simple illustration of the ill-posed nature of the task is that for any fixed solution $\bm{\tilde{I}}, \bm{\tilde{K}}_g$ of Eq.~\ref{equ:main_blurring_model}, the family of $\bm{\tilde{I}} \cdot \lambda, \frac{\bm{\tilde{K}}_g}{\lambda}$ is also a valid solution, which is referred to as the scaling ambiguity.

Blind deblurring methods can be divided into three categories, based on the approach for obtaining the sharp image (estimation of the latent image in Eq.~\ref{equ:main_blurring_model}). Each category includes an extensive literature, hence 
only the most closely related to our work are summarised below. For a more thorough study, the interested reader is redirected to \cite{laicomparative}.

\textbf{Synthesis-based}: 
Instead of solving the optimisation, these methods typically include a heuristic for 'guessing' the blurry parts and 
then apply a synthesis-based replacement in the blurry regions.
The majority of those works (\cite{cho2012video, tan2016kernel}) implicitly assume 
that i)~there are multiple frames with approximately the same content and that ii)~there exists a 
sharp patch that matches in content the respective blurry one. These two strong assumptions 
combined with the poor texture in a face (which weakens the heuristic to detect 
sharp patches), result in not using synthesis-based methods for face deblurring. 

\textbf{Optimisation-based}:
Based on Eq.~\ref{equ:main_blurring_model} and assuming $\psi$ is the identity function, this class of methods formulates the problem as the minimisation of a cost function of the format 
\begin{equation}
\tilde{\bm{I}} = \argmin_{\bm{I}}(||\bm{I}_{bl} - \bm{I} * \bm{K}_g||_2^2 + f(\bm{I}_{bl})).
\label{equ:optimisation_based_cost}
\end{equation}
with $f(\bm{I}_{bl})$ a set of priors based on generic image statistics or domain-specific priors.
These methods are applied in a coarse-to-fine manner, while they estimate the (dense) kernel and 
then perform a non-blind deconvolution.

The estimation of the blur kernel $\bm{K}_g$ and the latent image $\bm{I}$ occur in an alternating manner, which might lead 
to a blurry $\tilde{\bm{I}}$ if a joint MAP (Maximum a posteriori) optimisation is followed 
(\cite{levin2009understanding}).
Levin \etal suggest instead to solve a MAP on the kernel with a gradient-based prior based natural image statistics.
More recently, Pan \etal in \cite{pan2014deblurring_text} apply an $\ell_0$ norm as a sparse 
prior on both the intensity values and the image gradient for deblurring text. 
HaCohen \etal in~\cite{hacohen2013deblurring} support that the gradient prior alone 
is not sufficient, and introduce a prior that locates dense correspondences 
of the blurry image with a similar sharp image, 
while they iteratively optimise over the correspondence, the kernel and the sharp image estimation. 
A strong requirement of their algorithm is the similar reference image, which is not always available. 
A generalisation of \cite{hacohen2013deblurring} is the work of \cite{pan2014deblurring}, 
which also requires an exemplar dataset to locate an image with a similar contour.
However, in \cite{pan2014deblurring} the authors restrict the task to face deblurring to profit from the shape structure. A search in a dataset with exemplar images is performed to locate an image with a similar contour as the test image. The gradient of the exemplar image provides the initial blind estimation iterations, which leads
to an improved performance. Unfortunately, the noisy contour matching process 
along with the obligatory presence of a similar contour in the
dataset limit the applications of this work.

Even though the optimisation-based methods have proven to work well with synthetic blurs, they do not generalise well in
real world blurred images (\cite{laicomparative}) due to the strong assumptions of invariance and the simplified 
format of $\psi$. Another common attribute of these methods is the iterative optimisation procedure; they are executed in a loop hundreds or even thousands of times to return a deblurred image, which classifies these methods as computationally intensive; some of them require hours for deblurring a single image (\cite{chakrabarti2016neural}). %

\textbf{Learning-based}:
With the resurrection of neural networks, few approaches for learning a network to perform deblurring have emerged. 
The experimental superiority of neural networks as function approximators consists a strong motivation for relying on neural networks for deblurring. The non-linear units allow us to model non-linear functions $\psi$ or spatially varying blur kernels. 
Obtaining a sharp image in this case is defined as a function 
$\tilde{\bm{I}} = \phi(\bm{I}_{bl}, \bm{p})$, with $\bm{p}$ denoting the hyper-parameters of the method. 

Some methods (\cite{hradivs2015convolutional}) learn straight away the function $\phi$ from the data, 
while others (\cite{sun2015learning, chakrabarti2016neural}) learn an estimate and perform 
non-blind de-blurring/refinement of the sharp image.
In~\cite{hradivs2015convolutional}, the regularised $\ell_2$ loss of an up to 15-layer CNN is minimised for text deblurring. Even though they report nice results, the text deblurring domain is a structured but limited class (the sharp text can be represented as a sequence of binary intensity values). They argue that the performance can be mainly 
attributed to the network that modelled well the text prior, 
hence it is questionable whether this would work in more complex object types.
Sun \etal in~\cite{sun2015learning}, learn a CNN to recognise few discretised motion kernels and 
then perform a non-blind deconvolution in a dense motion field estimate. 

Our method belongs in the learning-based category, specifically the methods that learn $\phi$ from the data. The combination of such a learning method with weak supervision through landmark localisation has not been performed before for deblurring. 

\section{Method}
\label{sec:deblurring_method}
In this Section, we portray our learning-based method for face deblurring. 
We develop our way for providing the required input for the network (pairs of blurry/sharp images). Sequentially we introduce the deep architecture that we employed, along with the pre-processing step to take advantage of the facial structure through landmark localisation. Finally, we refer to the inference steps for an unseen image.

\subsection{Notation}
A sparse shape of $n$ fiducial (landmark) points is denoted as $\bm{l}$ for 
the image $\bm{I}$ with 
$\bm{l}\,= [[\bm{\ell}_1]^T,  [\bm{\ell}_2]^T, ..., [\bm{\ell}_n]^T]^T$, 
with $\bm{\ell}_j\,=\,[x_j, y_j]^T,  j\,\in [1, n], x_j, y_j \in \mathbb{R}$ the Cartesian coordinates 
of the j\textsuperscript{th} point.
When referring to a random image $\bm{I}$, we hypothesise that 
$\bm{I}$ contains a human face, of which the facial sparse shape $\bm{l}$ is available.

\subsection{Training pair creation}
The dominant way to discriminatively train a network is by feeding pairs of input and label samples; the labels are used to compute the error and improve the network performance. In our case the input is the blurry image, the label is the corresponding sharp image. Obtaining real world blurred images with a dense correspondence with a similar sharp image is not trivial, especially if thousands such pairs are required to train a deep network. Hence, following similar methods (\cite{sun2015learning, chakrabarti2016neural}) we resort to simulating the blur from sharp images.

A synthetically blurred image $\bm{I}_{bl}$ is generated by convolving the original sharp image $\bm{I}$ with a blur kernel (simulating Eq.~\ref{equ:main_blurring_model}). A unique blur kernel is created for every input image to allow for the maximum variation in the number of blur kernels that have emerged during the training.
The blur kernel is chosen arbitrarily in each step between a Gaussian blur kernel and a motion blur kernel, both with varying deviation and spatial support.

\begin{figure*}[!htb]
    \centering
    \includegraphics[width=0.161\linewidth]{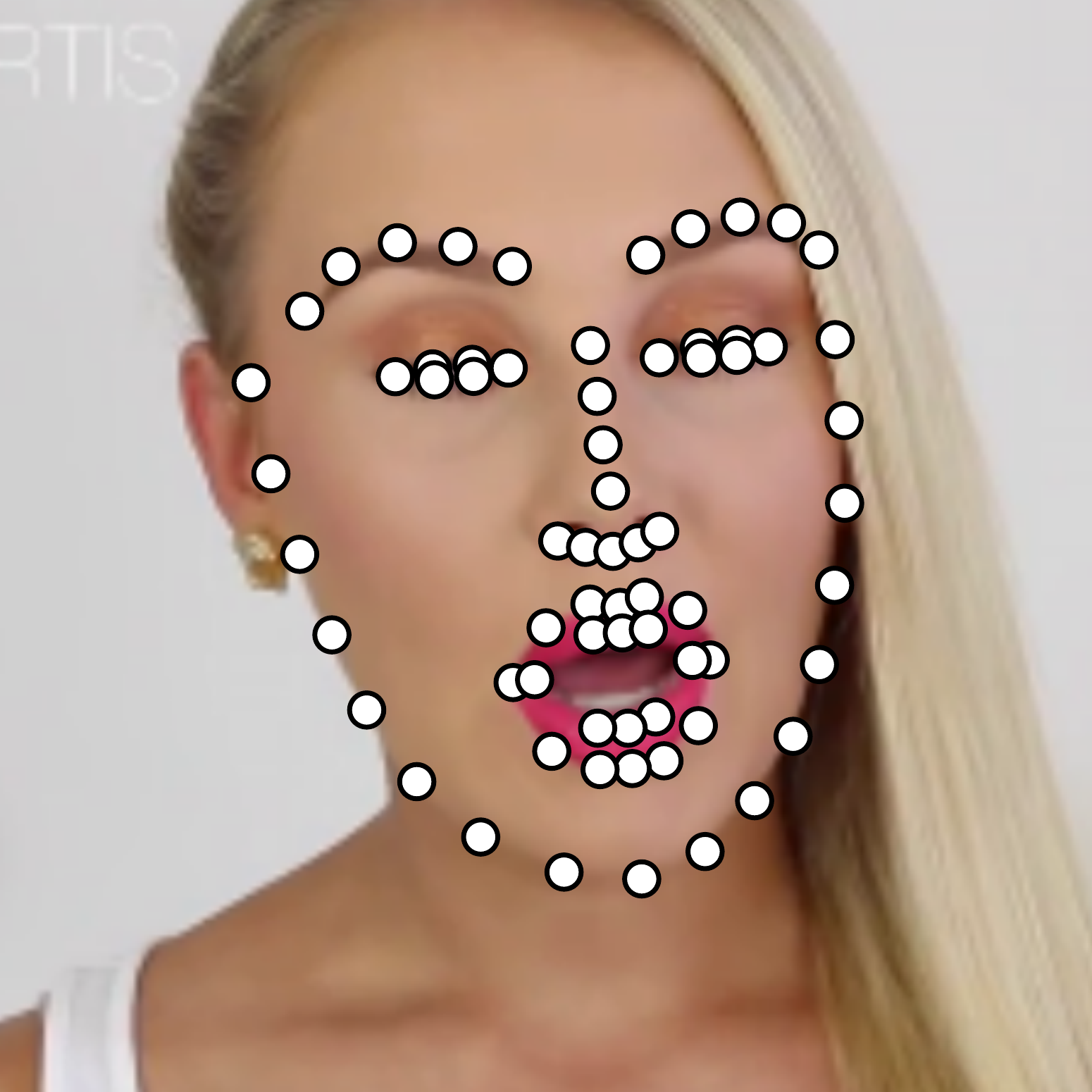} 
    \includegraphics[width=0.161\linewidth]{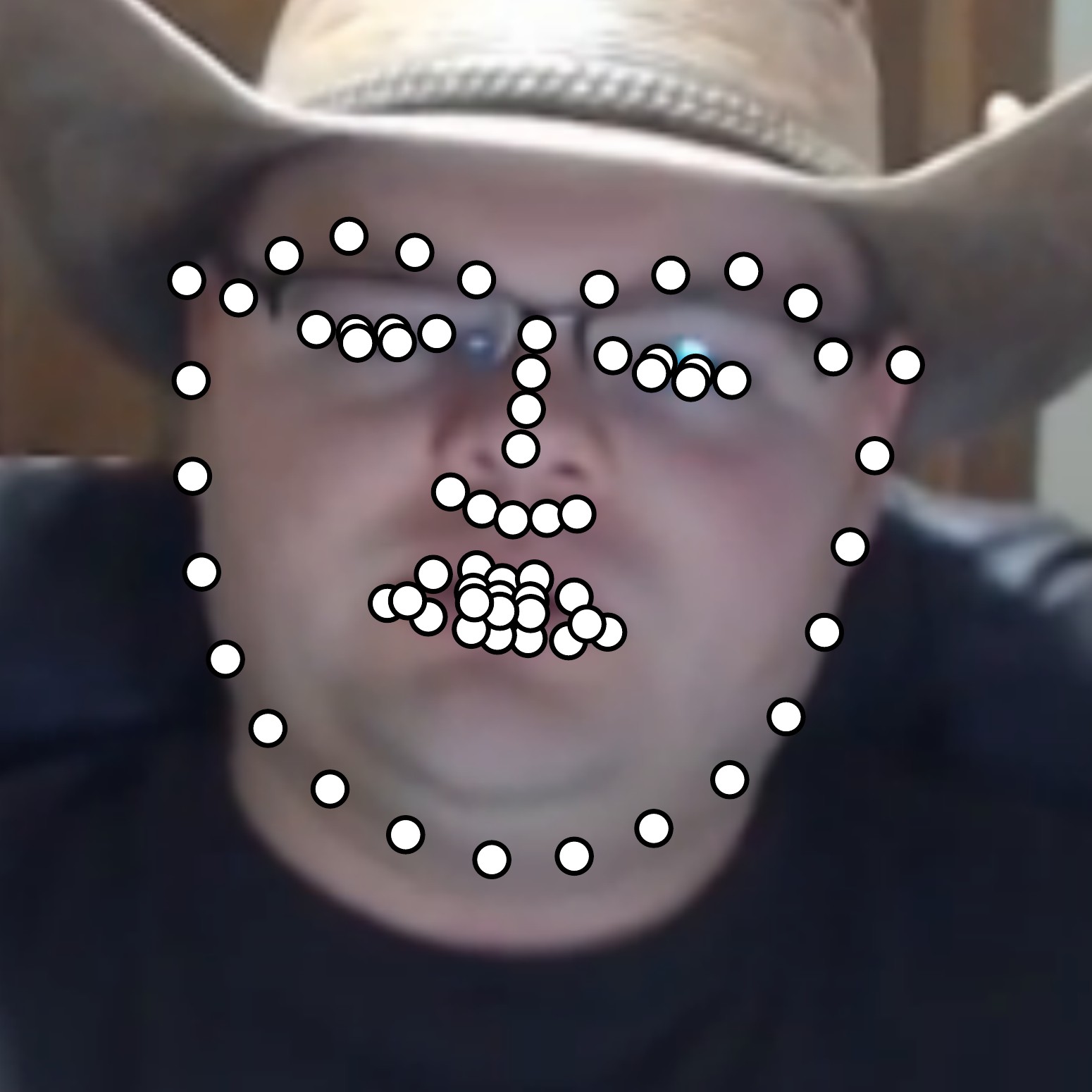} 
    \includegraphics[width=0.161\linewidth]{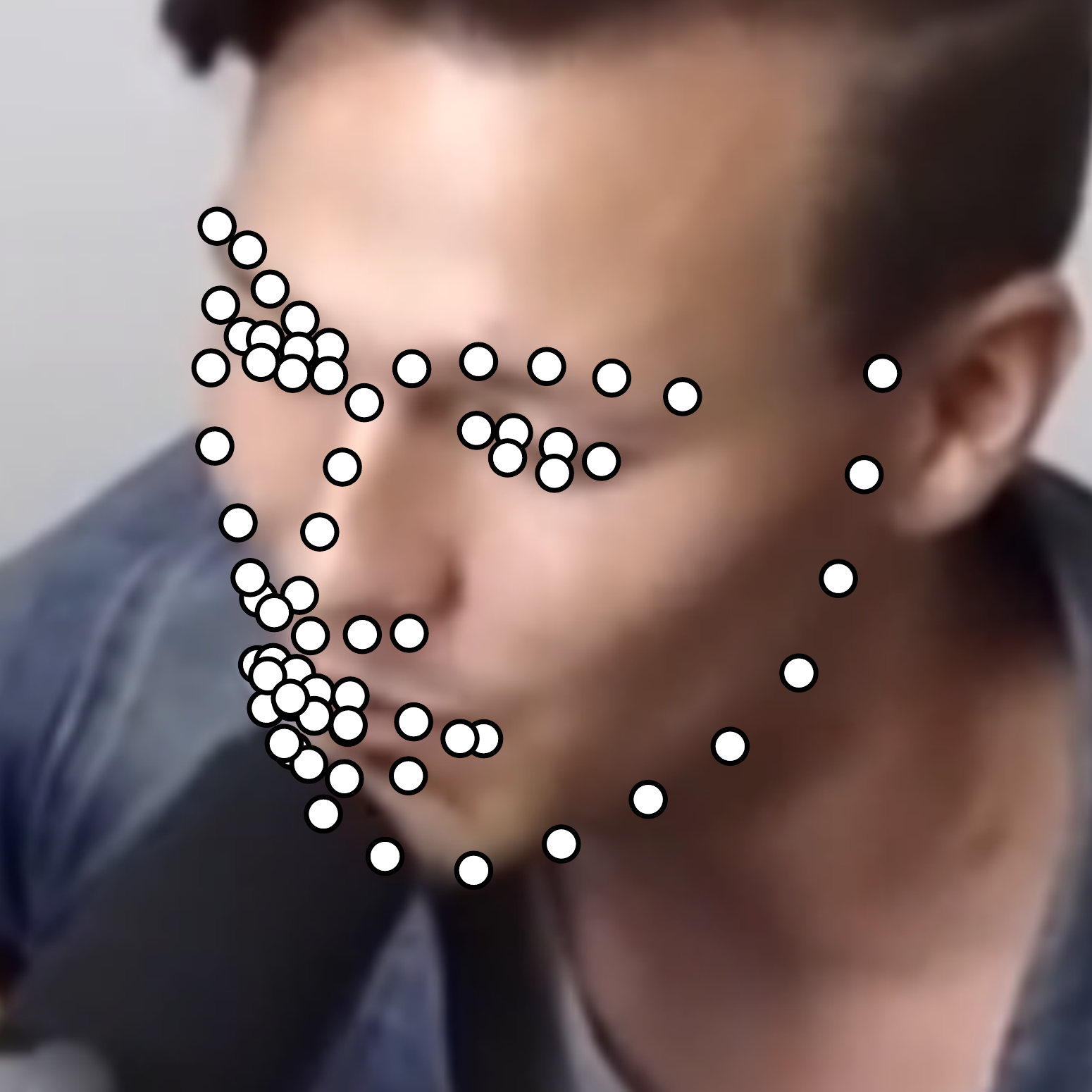} 
    \includegraphics[width=0.161\linewidth]{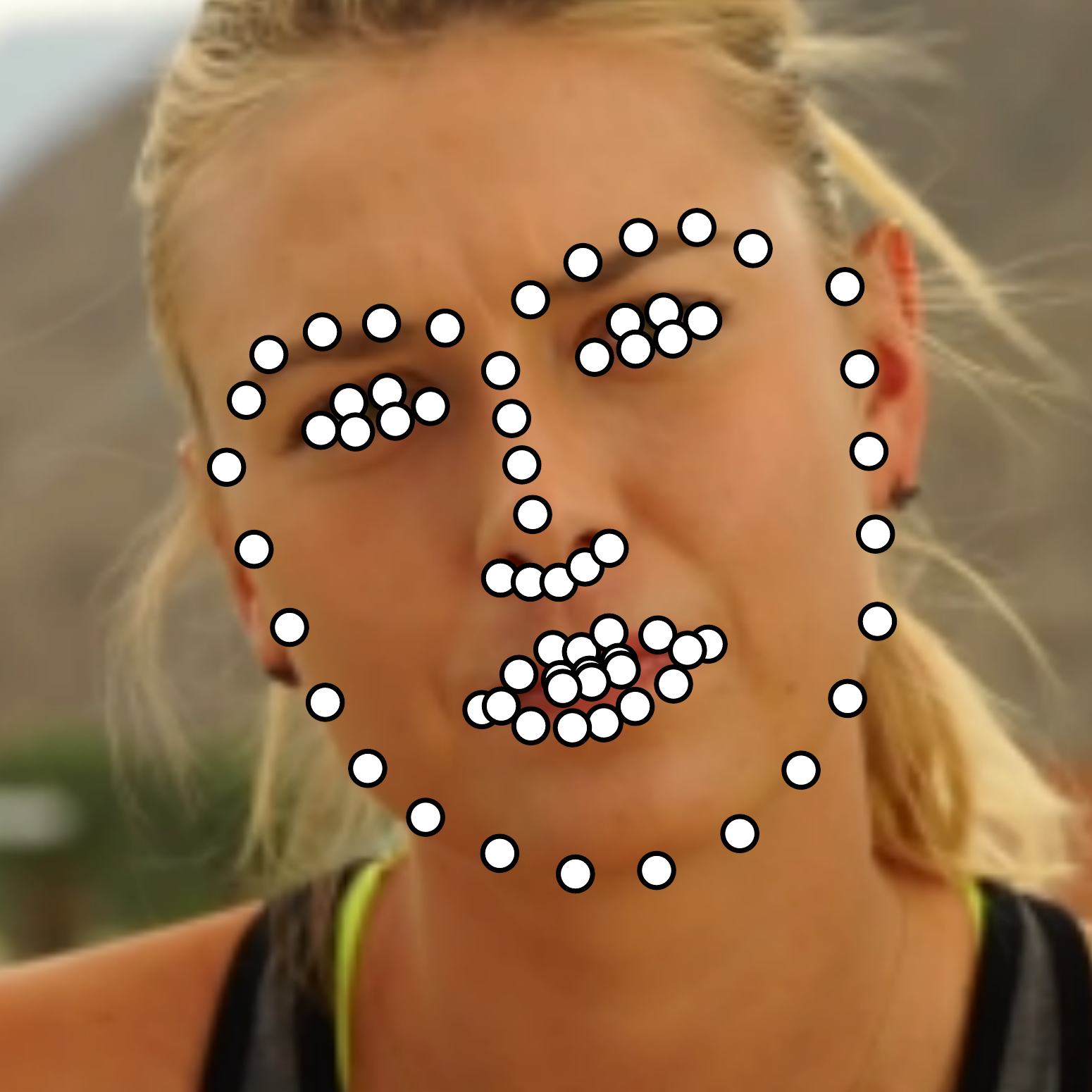} 
    \includegraphics[width=0.161\linewidth]{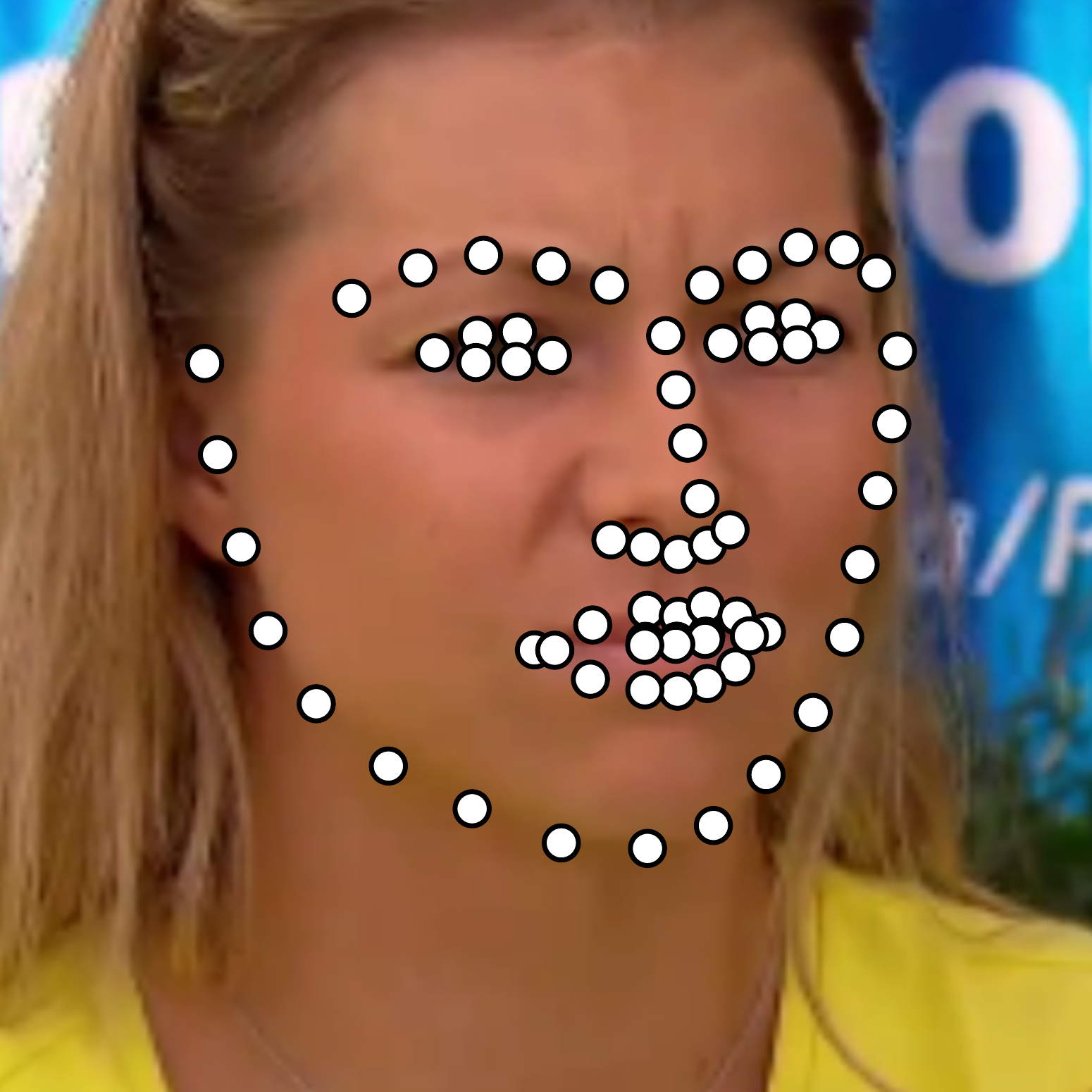} 
    \includegraphics[width=0.161\linewidth]{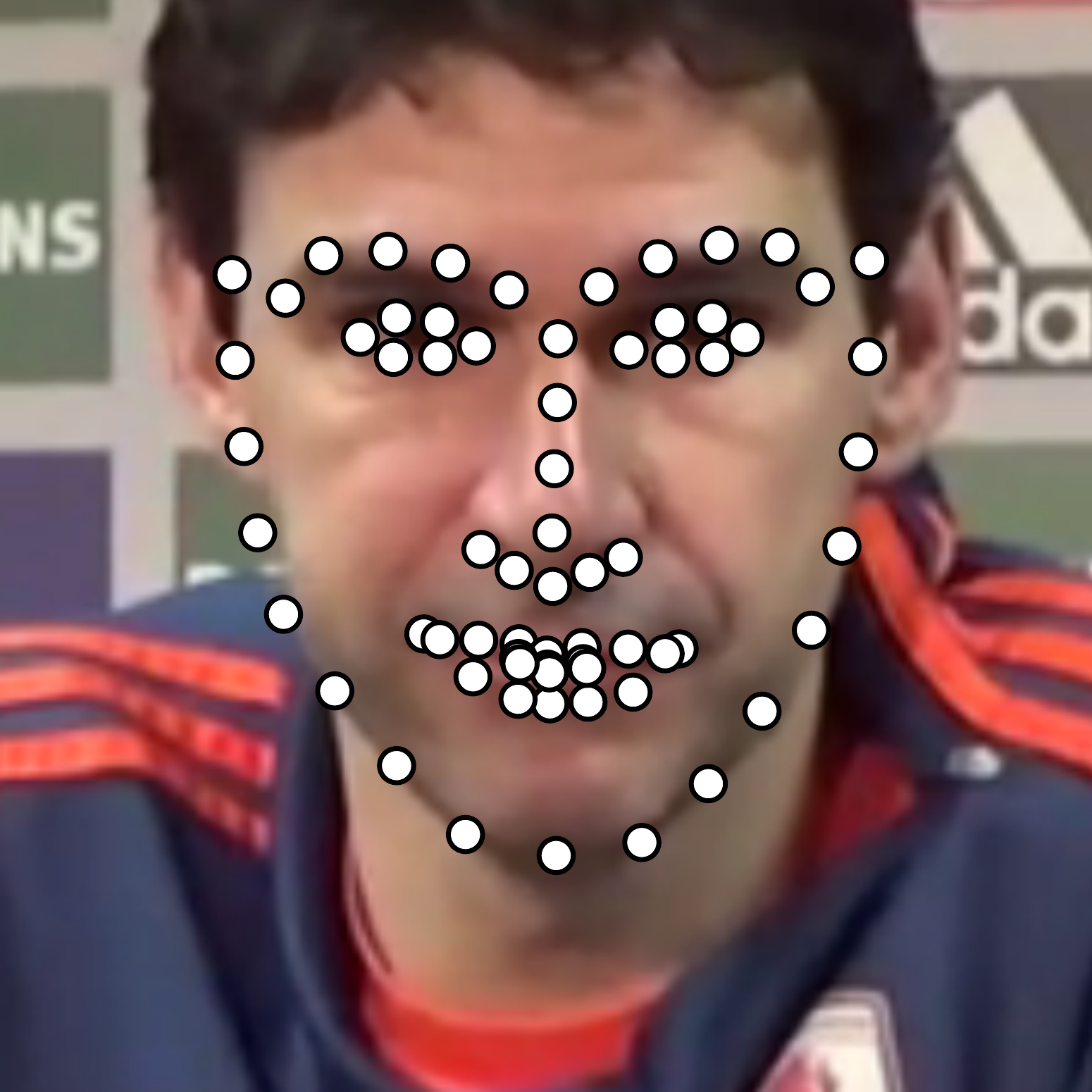} 
    \caption{Sample frames from the $2MF^2$ dataset along with the sparse shape denoted with white dots.}
    \label{fig:deblurring_samples_2mf2}
\end{figure*}

\subsection{Network}
Following the reasoning of \cite{hradivs2015convolutional} that demonstrated the success of a 15-layer CNN for the simpler task of text deblurring, we employ a network with several convolutional layers to allow a richer representation to be learnt. A modified version of the residual network (ResNet) architecture of \cite{he2015deep} is used as the learning component in our method.  ResNet consists of a number of `blocks'; each  `block' is a sequence of convolutional layers, 
followed by Rectified Linear Units, with identity connections connecting the blocks.
This simple architecture has demonstrated state-of-the-art performance in several tasks, while there is an effort to establish their dominance from a theoretical perspective (\cite{hardt2016identity}).

We modify the original ResNet by disabling all the max pooling operations, while skip connections (\cite{hariharan2015hypercolumns}) are added in the $2^{nd}$ and $3^{rd}$ ResNet blocks. A batch normalisation is added in every skip connection to ensure a common scale; a linear mapping is learnt from the high-dimensional space of the connections to the low dimensional space of the output image shape. 
The huber loss of \cite{huber1973robust} is utilised for our loss function. This is a continuous and differentiable function with
\begin{equation}
    L_h(\bm{x}) = \begin{cases} 
                    ||\bm{x}||_1 - 0.5 & ||\bm{x}||_1 > 1\\
                    0.5||\bm{x}||_2^2 & otherwise
                \end{cases}
\label{equ:huber_loss}
\end{equation}
Namely, the loss function of our network is:
\begin{equation}
    L = L_h(\phi(\bm{I}_{bl}) - \bm{I})
\label{equ:network_loss}
\end{equation}

\subsection{Inference}
The single input during inference is the blurry image $\bm{I}_{bl}$, \ie no latent image as ground-truth is required. The blurry image is pre-processed to obtain the appropriate region of the image to be fed into the network. Concretely, an off-the-shelf face detector is employed to acquire the bounding box; the landmarks are localised through a localisation technique. The image is rescaled based on the size of the landmarks, while a rectangular area around the face (landmarks) is cropped, which is the area that is fed into the network (only the feed-forward part of which is required). 

Among the most successful face detectors is the deformable part models (DPM) detector~\cite{felzenszwalb2010object, mathias2014face}. DPM learn a mixture of models which aim to detect faces in different poses. Each model implicitly considers some parts which are allowed to deform with a quadratic cost. The cost function of DPM contains an appearance (unary) term along with a pairwise (deformation) term plus a bias, all of which are learned with a discriminative training procedure. 
The crude bounding box of the DPM consists the initialisation of a landmark localisation technique~\cite{zhu2015face, kazemi2014one, chrysos2016comprehensive}. Both techniques~\cite{zhu2015face, kazemi2014one} belong to the regression based discriminative methods for landmark localisation. These methods learn to regress from the pixel intensities (with the former extracting hand-crafted SIFT features, while the latter of Kazemi \etal rely on data driven learned features) to the sparse shape coordinates. Both methods have proven very accurate in a number of benchmarks~\cite{sagonas2015300, chrysos2016comprehensive}, hence we adopt the method of Kazemi \etal due to a publicly available fast implementation~\cite{king2009dlib}.

\section{Data mining}
\label{sec:deblurring_dataset}

In this Section, we describe our method for mining frames from videos in a semi-supervised manner.
A number of videos are crawled using the API's of web sources, \eg Youtube; each video consists of few thousand frames and is analysed independently to determine the frames, if any, that are appropriate for the task. In our case, we aim at utilising the videos with dynamically moving faces. We defined the following three requirements for a video to be included in the training: 
\begin{enumerate}
	\item a face is present in each frame,
	\item the face is not completely static throughout the video,
	\item the video includes real world images, not synthetically generated ones.
\end{enumerate}
To that end, we have devised an efficient, automatic framework to perform this task; the steps are summarised in Alg.~\ref{alg:deblurring_framework_alg}.

The face detector of \cite{felzenszwalb2010object, mathias2014face} is applied to the first frame of the video. If there is no detection, the video is discarded, otherwise the bounding box obtained initialises a model-free tracker. Given the state of the first frame, a model-free tracker determines the state of the subsequent frames, while no prior information about the object is provided. The tracker should adapt to any appearance, deformation changes, which constitutes a very challenging task, thus an immense amount of diverse techniques has been proposed. In our work, we utilise the SRDCF tracker~\cite{danelljan2015learning}, which provides a decent trade-off of accurate deformable tracking quality and computation complexity~\cite{chrysos2016comprehensive}.

Even though SRDCF is robust to a wide range of variations, an additional criterion of overlap per frame with the bounding box of the DPM detector is performed. Specifically, we require the bounding boxes of the tracker and the detector to have at least a 50\% overlap (intersection over union overlap) in half of the frames, otherwise the clip is discarded. Subsequently, the landmark localisation technique of \cite{zhu2015face} is employed to obtain the sparse shape for each face. Due to the object-agnostic nature of the model-free tracker, we eliminate the few erroneous fittings by learning a statistical function $f_{cl}$. We utilise a linear patch-based SVM~\cite{cortes1995support} as the classifier $f_{cl}(\bm{I}, \bm{l})$ which accepts a frame $\bm{I}$ along with the respective fitting $\bm{l}$ and returns a binary decision on whether this is an acceptable fitting. The classifier fulfils the first requirement for every frame, \ie that a face is present. 

The requirement of non-static faces is fulfilled by computing the optical flow~\cite{farneback2003two} in the accepted frames and requiring that there is at least a pixel movement from frame to frame. If the average movement per pixel is above a threshold, the video is discarded. %

This framework can be adapted for different type of objects with two minor modifications. The modifications are: (i)~the face detection module, which can be trivially replaced by a generic detector like \cite{girshick2015fast}, (ii)~the classifier module for the removal of erroneous fittings, which should be trained for the task, \eg to accept the whole bounding box instead of the patch-based SVM utilising the landmarks. 

\begin{algorithm}[h]
	\DontPrintSemicolon
	\SetKwInOut{Input}{Input}
	\SetKwInOut{Output}{Output}
	\SetKwInOut{Init}{Initialize}
	
	\Input{Video frames $\bm{V} = [\bm{I}^{(1)}, \bm{I}^{(2)}, \ldots, \bm{I}^{(M)}]$}
	\Output{Accepted frames $\bm{F}$, Landmarks $\bm{L}$}
	\Init{$\bm{F} = [], \bm{L} = [], cnt\_over = 0$}
	
	\tcc{detection in the first frame.}
	$\bm{faces}$ = face-detection($\bm{I}^{(1)}$)\;
	\If{length($\bm{faces}$) $== 0$}
	{
		return $\bm{F}, \bm{L}$ 
	}
	\tcc{bb: tracker's bounding box.}
	$\bm{bb}$ = $\bm{faces}[0]$\;
	
	\tcc{main tracking loop.}
	\For{idx = 1 to M}
	{
		$\bm{faces}$ = face-detection($\bm{I}^{(idx)}$)\;
		$\bm{bb}$ = track($\bm{I}^{(idx)}$, $\bm{bb}$)\;
		\If{length($\bm{faces}$) $> 0$ and compute\_overlap($\bm{faces}[0]$, $\bm{bb}$) $ > 0.5$}
		{
			$cnt\_over += 1$\;
		}
		$\bm{l}^{(idx)} = landmark\_localisation(\bm{I}^{(idx)}, \bm{bb})$\;
		\tcc{$f_{cl}$: classifier to reject the erroneous fittings.}
		accept\_fitting = $f_{cl}(\bm{I}^{(idx)}, \bm{l}^{(idx)})$\;
		\If{accept\_fitting}
		{
			append($\bm{F}, \bm{I}^{(idx)}$)\;
			append($\bm{L}, \bm{l}^{(idx)}$)\;
		}
	}
	\If{$cnt\_over < M / 2$}
	{
		return $[], []$
	}
	return $\bm{F}, \bm{L}$ 
\caption{The automatic framework as introduced in Sec.~\ref{sec:deblurring_dataset} to create the $2MF^2$ dataset.}
\label{alg:deblurring_framework_alg}
\end{algorithm}

The aforementioned framework was utilised to create $2MF^2$ (2 million frames of faces). $2MF^2$ consists of 1150 videos, with 2,1 million accepted frames that contain a human face. Exemplar frames of few videos are visualised in Fig.~\ref{fig:deblurring_samples_2mf2}, while an accompanying video depicting accepted frames along with the derived sparse shape can be found in \url{https://youtu.be/Mz0918XdDew}.   
\section{Experiments}
\label{sec:deblurring_experiments}
In this Section we develop few implementation details, summarise a validation experiment for our method with a simple Gaussian blur, compare with the state-of-the-art methods for deblurring in two different scenarios, which include motion blur and real world blurred images.

\subsection{Implementation details}
\label{ssec:deblurring_implementation}

The network was implemented in Tensorflow~\cite{tensorflow2015-whitepaper} using the Python API; 
the pre-trained weights of the network were obtained from the original ResNet paper~\cite{he2015deep}, 
while the majority of the rest functionality was provided by the Menpo project~\cite{menpo14}. 

\begin{figure}[!htb]
    \centering
    \includegraphics[width=0.238\linewidth]{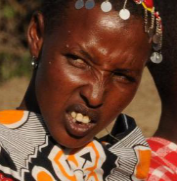} 
    \includegraphics[width=0.238\linewidth]{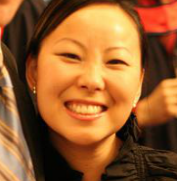} 
    \includegraphics[width=0.238\linewidth]{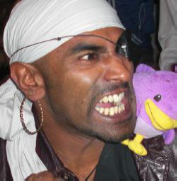} 
    \includegraphics[width=0.238\linewidth]{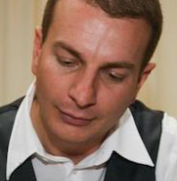} \\

    \includegraphics[width=0.238\linewidth]{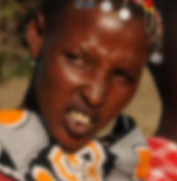} 
    \includegraphics[width=0.238\linewidth]{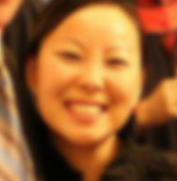} 
    \includegraphics[width=0.238\linewidth]{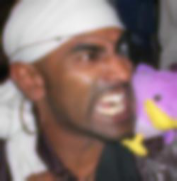} 
    \includegraphics[width=0.238\linewidth]{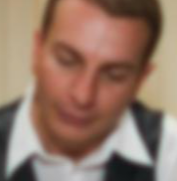} \\

    \includegraphics[width=0.238\linewidth]{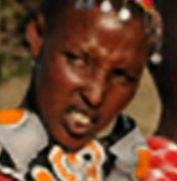} 
    \includegraphics[width=0.238\linewidth]{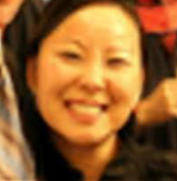} 
    \includegraphics[width=0.238\linewidth]{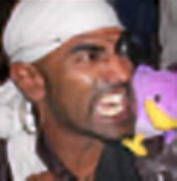} 
    \includegraphics[width=0.238\linewidth]{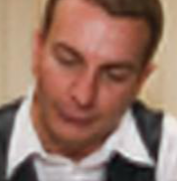} \\

    \caption{(Preferably viewed in colour) Visual results for the self evaluation experiment. On the first row is the original image, on the middle the blurred image and the last consists of the outputs of our network.}
    \label{fig:deblurring_self_evaluation_qualitative}
\end{figure}
The shapes of the public datasets with 68 facial points mark-up annotation, 
\ie IBUG~\cite{sagonas2013300}, HELEN~\cite{le2012interactive}, 
LFPW~\cite{belhumeur2013localizing} and the 300W~\cite{sagonas2015300} were utilised 
a) for training the classifier of Sec.\ref{sec:deblurring_dataset}, b) as additional input to the network for training. 
Few images with severe distortions were excluded from the training set; 
the frames of $2MF^2$ were sub-sampled and one every $2^{nd}$ frame was used for the training.
The training steps of the classifier were the following: (a) The positive training samples were extracted from the 300W trainset; perturbed versions of the annotations of those images along with selected images of Pascal dataset~\cite{everingham2010pascal} were used for mining the negative samples. (b) A fixed size patch was extracted from each positive sample around each of the $n$ landmark points; SIFT~\cite{lowe2004distinctive} were computed per patch. For each negative sample a random perturbation of the ground truth points was performed to create an erroneous fitting prior to extracting the patches. (c) A linear SVM was trained, with its hyper-parameters cross-validated in withheld validation frames.

For training our network, we used a mini-batch size of 16; SGD with an exponentially decreasing learning rate (initial value of 0.0003), and decreasing by a factor of 0.5 every 15k iterations. 
The final training consisted of 70k iterations and was completed in a single-core GPU machine. It should be noted that each frame was loaded only once in the network, to avoid over-fitting the training data. Our method functions at 6 fps in a GPU Titan X machine. 

\begin{figure*}
    \centering
    \includegraphics[width=0.106\linewidth]{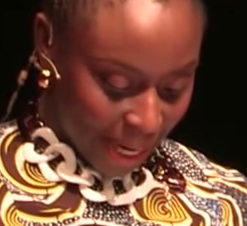} \hspace{-1.1mm}
    \includegraphics[width=0.106\linewidth]{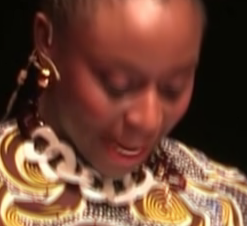} \hspace{-1.1mm}
    \includegraphics[width=0.106\linewidth]{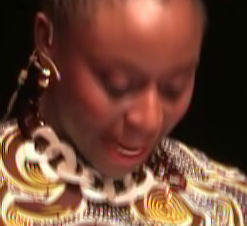} \hspace{-1.1mm}
    \includegraphics[width=0.106\linewidth]{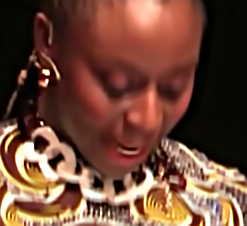} \hspace{-1.1mm}
    \includegraphics[width=0.106\linewidth]{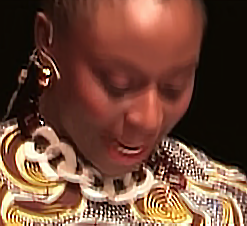} \hspace{-1.1mm}
    \includegraphics[width=0.106\linewidth]{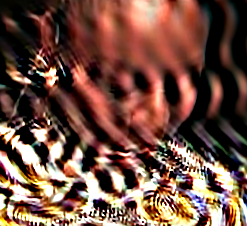} \hspace{-1.1mm}
    \includegraphics[width=0.106\linewidth]{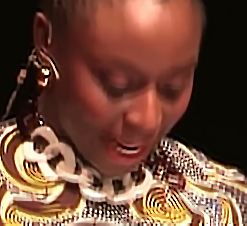} \hspace{-1.1mm}
    \includegraphics[width=0.106\linewidth]{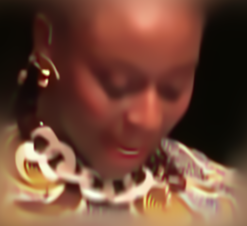} \hspace{-1.1mm}
    \includegraphics[width=0.106\linewidth]{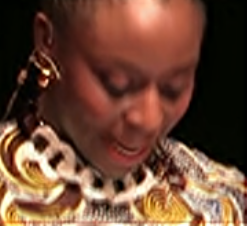} \hspace{-1.1mm}\\

    \includegraphics[width=0.106\linewidth]{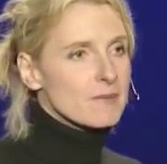} \hspace{-1.1mm}
    \includegraphics[width=0.106\linewidth]{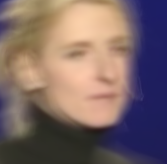} \hspace{-1.1mm}
    \includegraphics[width=0.106\linewidth]{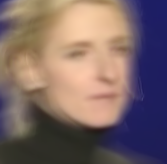} \hspace{-1.1mm}
    \includegraphics[width=0.106\linewidth]{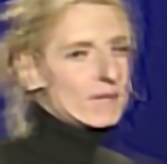} \hspace{-1.1mm}
    \includegraphics[width=0.106\linewidth]{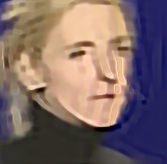} \hspace{-1.1mm}
    \includegraphics[width=0.106\linewidth]{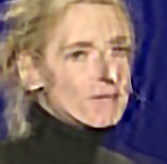} \hspace{-1.1mm}
    \includegraphics[width=0.106\linewidth]{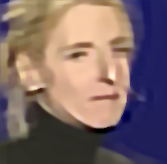} \hspace{-1.1mm}
    \includegraphics[width=0.106\linewidth]{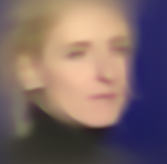} \hspace{-1.1mm}
    \includegraphics[width=0.106\linewidth]{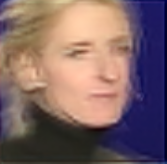} \hspace{-1.1mm}\\ 
    
    \includegraphics[width=0.106\linewidth]{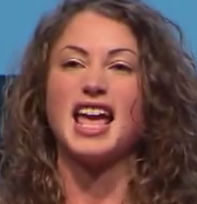} \hspace{-1.1mm}
    \includegraphics[width=0.106\linewidth]{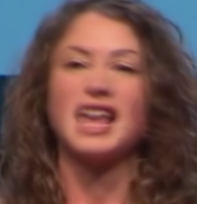} \hspace{-1.1mm}
    \includegraphics[width=0.106\linewidth]{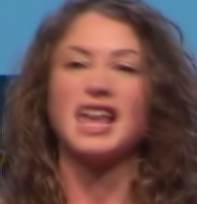} \hspace{-1.1mm}
    \includegraphics[width=0.106\linewidth]{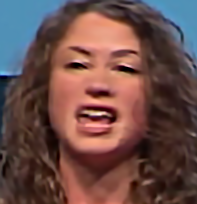} \hspace{-1.1mm}
    \includegraphics[width=0.106\linewidth]{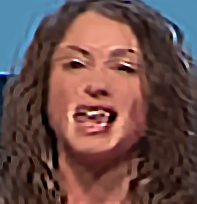} \hspace{-1.1mm}
    \includegraphics[width=0.106\linewidth]{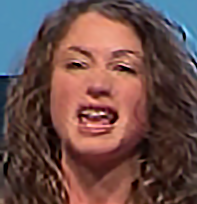} \hspace{-1.1mm}
    \includegraphics[width=0.106\linewidth]{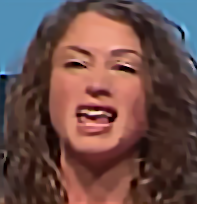} \hspace{-1.1mm}
    \includegraphics[width=0.106\linewidth]{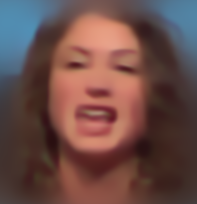} \hspace{-1.1mm}
    \includegraphics[width=0.106\linewidth]{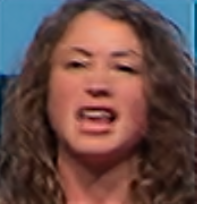} \hspace{-1.1mm}\\
    
    \vspace{-3mm} \hspace{-0.3mm}
    \subfloat[][GT]{\includegraphics[width=0.1043\linewidth]{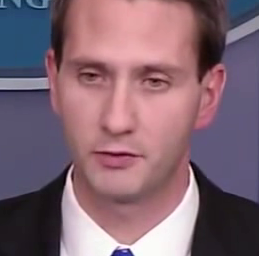}} \hspace{0.047mm}
    \subfloat[][Blurred]{\includegraphics[width=0.1043\linewidth]{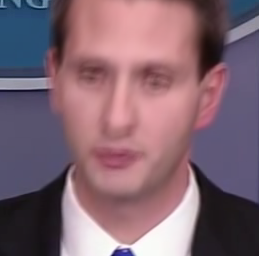}} \hspace{0.047mm}
    \subfloat[][\cite{babacan2012bayesian}]{\includegraphics[width=0.1043\linewidth]{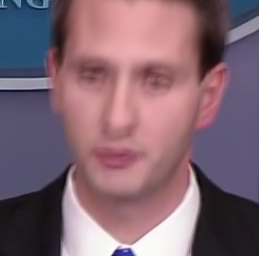}} \hspace{0.047mm}
    \subfloat[][\cite{zhang2013multi}]{\includegraphics[width=0.1043\linewidth]{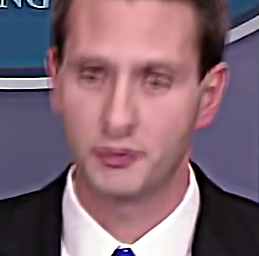}} \hspace{0.047mm}
    \subfloat[][\cite{pan2014deblurring}]{\includegraphics[width=0.1043\linewidth]{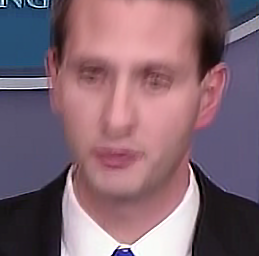}} \hspace{0.047mm}
    \subfloat[][\cite{pan2014deblurring_text}]{\includegraphics[width=0.1043\linewidth]{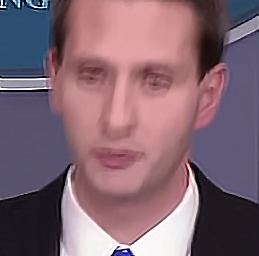}} \hspace{0.047mm}
    \subfloat[][\cite{panblind}]{\includegraphics[width=0.1043\linewidth]{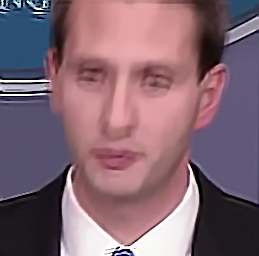}} \hspace{0.047mm}
    \subfloat[][\cite{chakrabarti2016neural}]{\includegraphics[width=0.1043\linewidth]{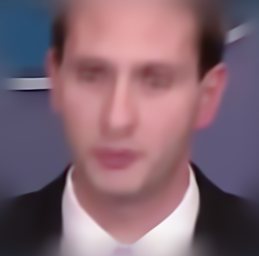}} \hspace{0.047mm}
    \subfloat[][Proposed]{\includegraphics[width=0.1043\linewidth]{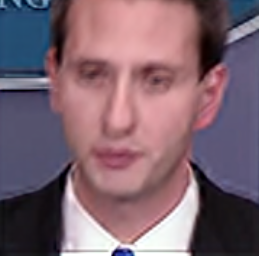}} \hspace{0.047mm}
    
    \caption{ (Preferably viewed in colour) Qualitative results for the simulated blur experiment.}
    \label{fig:deblurring_simulated_blur_qualitative}
\end{figure*}

\subsection{Self evaluation}
\label{ssec:deblurring_self_evaluation}
Seventy images of AFLW~\cite{kostinger2011annotated} were used to validate the outcome of the network. The images were synthetically blurred with Gaussian noise, while the standard visual quality metrics of PSNR and SSIM~\cite{wang2004image} were employed to compare the blurred images with the outputs of our network. The quality metrics are reported in Tab.~\ref{tbl:deblurring_quality_metrics_self_evaluation} and few indicative images are visualised in Fig.~\ref{fig:deblurring_self_evaluation_qualitative}. Both the qualitative and quantiative metrics indicate that the method indeed works well under Gaussian blur.

\begin{table}
    \begin{tabular}{|c|c|c|}
        \hline
        Image type/Quality metric & PSNR & SSIM \\ \hline
        Blurred                   & 21.84 & 0.52 \\ 
        Deblurred                 & 22.42 & 0.57 \\
        \hline
    \end{tabular}
    \caption{Image quality metrics for the validation of the network's outputs.}
    \label{tbl:deblurring_quality_metrics_self_evaluation}
\end{table}

\begin{figure*}
    \centering
    \includegraphics[width=0.119\linewidth]{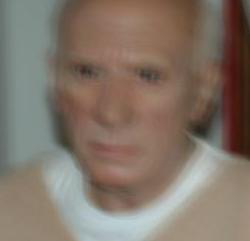} 
    \includegraphics[width=0.119\linewidth]{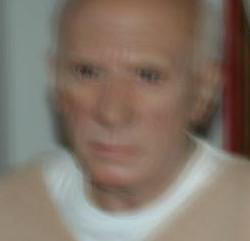} 
    \includegraphics[width=0.119\linewidth]{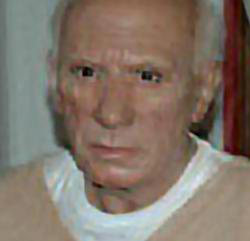} 
    \includegraphics[width=0.119\linewidth]{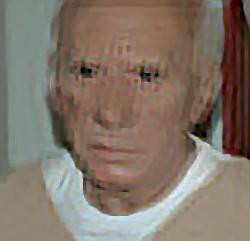} 
    \includegraphics[width=0.119\linewidth]{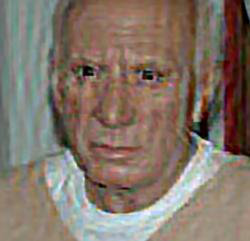} 
    \includegraphics[width=0.119\linewidth]{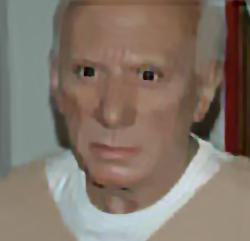} 
    \includegraphics[width=0.119\linewidth]{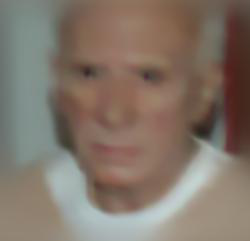} 
    \includegraphics[width=0.119\linewidth]{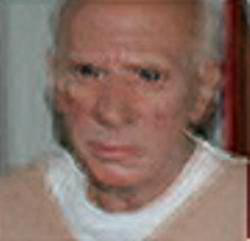} 
\\
    \includegraphics[width=0.119\linewidth]{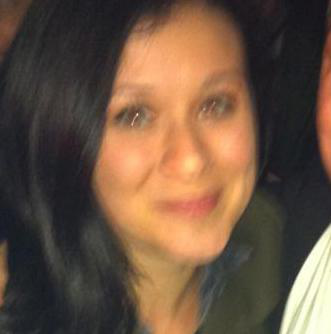} 
    \includegraphics[width=0.119\linewidth]{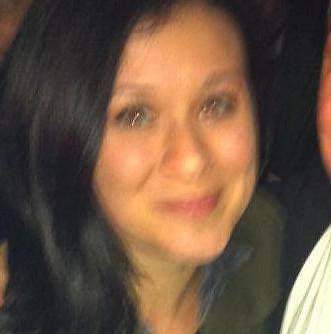} 
    \includegraphics[width=0.119\linewidth]{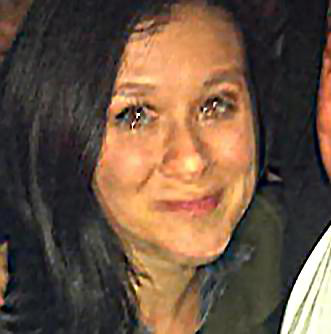} 
    \includegraphics[width=0.119\linewidth]{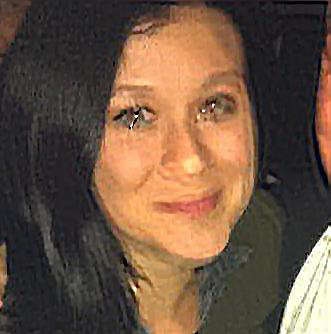} 
    \includegraphics[width=0.119\linewidth]{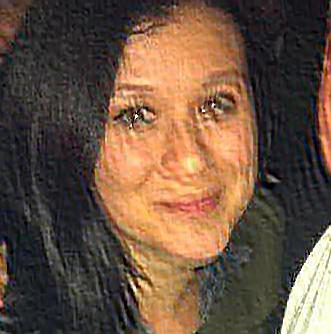} 
    \includegraphics[width=0.119\linewidth]{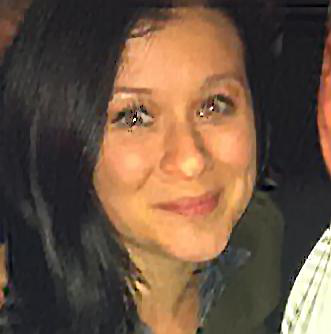} 
    \includegraphics[width=0.119\linewidth]{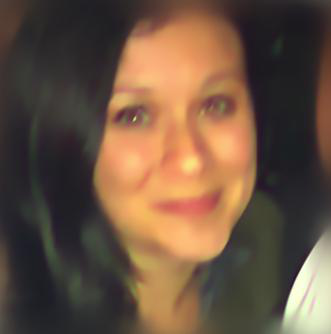} 
    \includegraphics[width=0.119\linewidth]{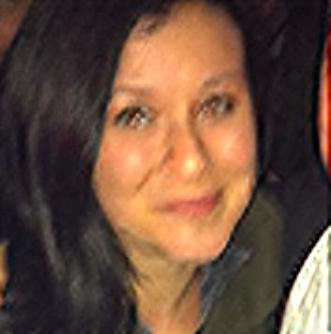} 
\\ \vspace{-3mm} \hspace{-0.2mm}
    \subfloat[][Blurred image]{\includegraphics[width=0.119\linewidth]{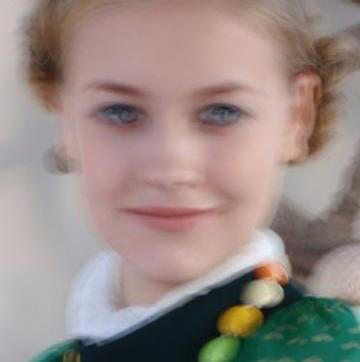}} \hspace{0.06mm}
    \subfloat[][\cite{babacan2012bayesian}]{\includegraphics[width=0.119\linewidth]{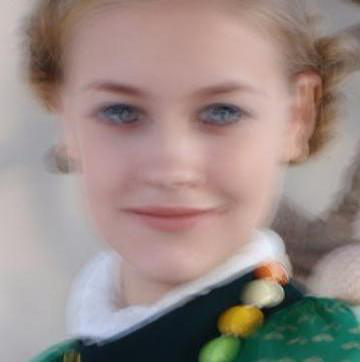}}\hspace{0.06mm}
    \subfloat[][\cite{zhang2013multi}]{\includegraphics[width=0.119\linewidth]{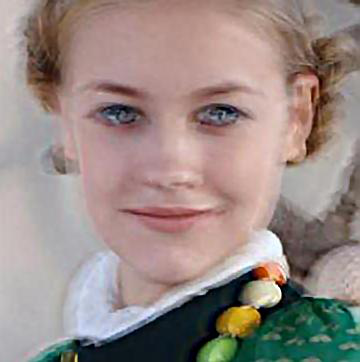}}\hspace{0.06mm}
    \subfloat[][\cite{pan2014deblurring_text}]{\includegraphics[width=0.119\linewidth]{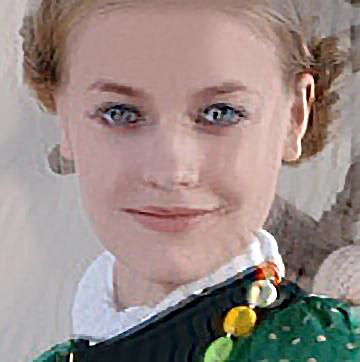}}\hspace{0.06mm}
    \subfloat[][\cite{pan2014deblurring}]{\includegraphics[width=0.119\linewidth]{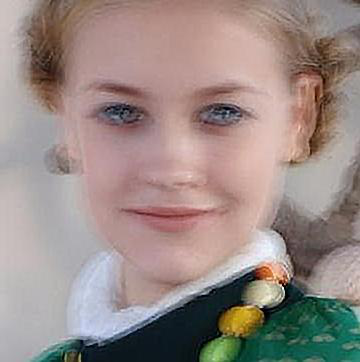}}\hspace{0.06mm}
    \subfloat[][\cite{panblind}]{\includegraphics[width=0.119\linewidth]{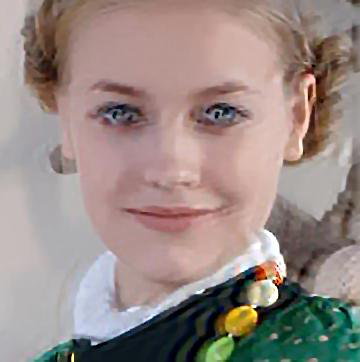}}\hspace{0.06mm}
    \subfloat[][\cite{chakrabarti2016neural}]{\includegraphics[width=0.119\linewidth]{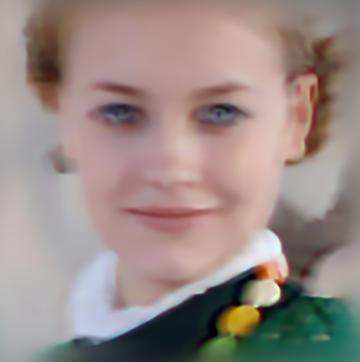}}\hspace{0.06mm}
    \subfloat[][Proposed]{\includegraphics[width=0.119\linewidth]{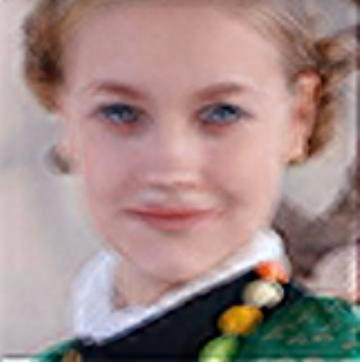}}
    \caption{(Preferably viewed in colour) Some images from the dataset of Lai \etal~\cite{laicomparative}. Notice that our method avoids the over-smoothing of other methods, \eg \cite{panblind}. Even though it deblurs the texture of the skin in a decent way, it sometimes suffers in localising the iris of the eye.}
    \label{fig:deblurring_real_world_compar_cvpr_study}
\end{figure*}

\begin{figure*}
	\centering 
    \includegraphics[width=0.192\linewidth]{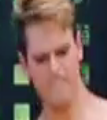} 
    \includegraphics[width=0.192\linewidth]{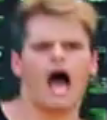} 
    \includegraphics[width=0.192\linewidth]{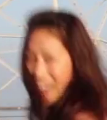} 
    \includegraphics[width=0.192\linewidth]{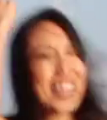} 
    \includegraphics[width=0.192\linewidth]{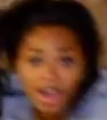} \\

    \includegraphics[width=0.192\linewidth]{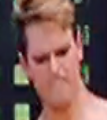} 
    \includegraphics[width=0.192\linewidth]{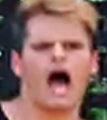} 
    \includegraphics[width=0.192\linewidth]{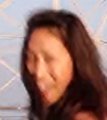} 
    \includegraphics[width=0.192\linewidth]{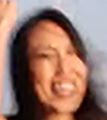} 
    \includegraphics[width=0.192\linewidth]{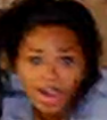}  
    \caption{(Preferably viewed in colour) Qualitative results in real world blurred images from arbitrary videos. On the top row, the original frame (there is no ground-truth available); on the second row the output of our method.}
    \label{fig:deblurring_qualitat_results}
\end{figure*}

\subsection{Simulated motion blur}
\label{ssec:deblurring_simulated_motion_blur}
To extend the simple Gaussian blur, an experiment that simulates motion blur observed in real world blurred images was conducted (this process of simulating the motion blur was only performed during testing). A set of sequential frames of a high frame rate video are averaged and simulate the movement of the person. The averaging creates the effect of a dynamic movement, while the middle frame of the averaging can be considered as the ground-truth frame. The edge cases of this simulation consist of (a)~no movement case, (b)~extreme movement case. The former case was avoided by considering the optical flow of each two sequential frames and ensuring there is at least some movement in the scene from frame to frame. For the latter case, the PSNR of the averaged frame was compared against the middle frame (ground-truth) and the frames below a threshold were discarded as too noisy.

In our experiment, four videos of the 300VW dataset~\cite{shen2015first} were utilised. All the videos of 300VW include a single person per video, while they are all over 25 fps. For each one of the videos employed, a different number of frames were averaged, ranging from 7 to 11 sequential frames. Also, the recent deblurring methods of Babacan \etal~\cite{babacan2012bayesian}, Zhang \etal~\cite{zhang2013multi}, Pan \etal~\cite{pan2014deblurring}, Pan \etal~\cite{pan2014deblurring_text}, Pan \etal~\cite{panblind} and Chakrabarti~\cite{chakrabarti2016neural} were also included in the experiment.
In Fig.~\ref{fig:deblurring_simulated_blur_qualitative}, the qualitative results of frames with simulated blur are visualised, while in Tab.~\ref{tbl:deblurring_quantitative_metrics_simulated_motion} the quantitative metrics are reported.

\begin{table}
    \begin{tabular}{|c|c|c|}
        \hline
        Image type/Quality metric                   & PSNR & SSIM \\ \hline
        Babacan \etal~\cite{babacan2012bayesian}    & 25.127 & 0.580 \\
        Zhang \etal~\cite{zhang2013multi}           & 23.303 & 0.521 \\
        Pan \etal~\cite{pan2014deblurring}          & 21.304 & 0.476 \\
        Pan \etal~\cite{pan2014deblurring_text}     & 22.492 & 0.473 \\
        Pan \etal~\cite{panblind}                   & 23.972 & 0.512 \\
        Chakrabarti~\cite{chakrabarti2016neural}    & 23.388 & 0.420 \\
        Proposed                                    & 23.950 & 0.558 \\
        \hline
    \end{tabular}
    \caption{Image quality metrics for the simulated motion blur experiment of Sec.~\ref{ssec:deblurring_simulated_motion_blur}.}
    \label{tbl:deblurring_quantitative_metrics_simulated_motion}
\end{table}

\subsection{Real world blurred images}
\label{ssec:deblurring_real_world_blurred_images_comp}
Providing a method that works for real world blurred images consists a strong motivation for our work. Unfortunately, comparing with real world blurred images comes at the cost of not having any ground-truth image\footnote{Capturing a real world blurred image and a sharp one with a dense correspondence requires an elaborate hardware/software setup. An approximation can be considered by capturing videos with a high frame rate camera (the middle frame can be used as the ground-truth), however this still does not guarantee the simulation to real world blurred image.}. Therefore, we opted to report the visual comparisons here. 

In Fig.~\ref{fig:deblurring_real_world_compar_cvpr_study}, the comparisons among different methods are provided for the facial images of Lai \etal~\cite{laicomparative}.
Additionally, to further emphasise the merits of the proposed method, we have gathered few images from internet sources in both indoors and outdoors scenes. The faces in those frames are of quite low-resolution, while there is rapid movement in the scne. The qualitative results are visualised in Fig.~\ref{fig:deblurring_qualitat_results}.
\section{Discussion and conclusions}
\label{sec:deblurring_conclusion}

In this work, we introduced a new method for deblurring facial images through inserting a weak supervision in the system, but not explicitly enforcing a strict alignment. The architecture that we have implemented is a modified version of the strong performing ResNet. We have also developed an automatic framework for large dataset creation  with off-the-shelf tools from the literature. Moreover, we have created $2MF^2$, a dataset that includes more than one thousand clips containing over two million frames of faces. The dataset was utilised to perform the training of our network. A number of experiments are conducted to validate the performance of our method and compare against the state-of-the-art deblurring methods.
\section{Acknowledgements}\label{sec:acks}

G. Chrysos was supported by EPSRC DTA award at Imperial College London. S. Zafeiriou was partially funded by the EPSRC Project EP/N007743/1 (FACER2VM).

{\small
\bibliographystyle{ieee}
\bibliography{egbib}
}

\end{document}